\documentclass[preprint,sort&compress,12pt]{elsarticle}

\usepackage{amssymb}
\usepackage{amsthm}
\usepackage{amsmath}
\usepackage{mathtools}
\usepackage{mathrsfs}
\usepackage{algorithm}
\usepackage[algo2e]{algorithm2e}
\usepackage{algpseudocode}
\usepackage{array}
\usepackage{multirow}
\usepackage{listings}
\usepackage{tabu}
\usepackage{enumerate}
\usepackage{lineno}
\usepackage{fullpage}
\usepackage{xcolor}
\usepackage[colorinlistoftodos]{todonotes}
\usepackage{bbm}
\usepackage[colorlinks=true]{hyperref}
\usepackage{url}
\usepackage{textcomp}
\usepackage{gensymb}
\usepackage{soul}
\usepackage{graphicx}
\usepackage{subfig}
\usepackage{float}
\usepackage{caption}
\usepackage{tikz}
\usetikzlibrary{shapes}

\usepackage{wrapfig}
\usepackage{enumitem}
\usepackage{nomencl}

\usepackage{changes}
\definechangesauthor[name={Reviewer 1}, color = blue]{R1}
\definechangesauthor[name={Reviewer 2}, color = red]{R2}
\definechangesauthor[name={Reviewer 1 \& 2}, color = blue]{All}
\definechangesauthor[name={Reviewer 3}, color = brown]{R3}
\definechangesauthor[name={Reviewer 4}, color = green]{R4}

\theoremstyle{definition}

\theoremstyle{remark}

\biboptions{numbers,comma,round,square}
\graphicspath{ {./figs/} }

\makenomenclature
\journal{Elsevier}
\begin{document}
\begin{frontmatter}


 \title{DiffHybrid-UQ: Uncertainty Quantification for Differentiable Hybrid Neural Modeling}



\author[ndAME]{Deepak Akhare}
\author[ndAME,Energy,ndCBE]{Tengfei Luo}
\author[ndAME,Energy,Lucy]{Jian-Xun Wang\corref{corxh}}


\address[ndAME]{Department of Aerospace and Mechanical Engineering, University of Notre Dame, Notre Dame, IN, USA}
\address[Energy]{Center for Sustainable Energy (ND Energy), University of Notre Dame, Notre Dame, IN, USA}
\address[ndCBE]{Department of Chemical and Biomolecular Engineering, University of Notre Dame, Notre Dame, IN, USA}
\address[Lucy]{Lucy Family Institute for Data \& Society, University of Notre Dame, Notre Dame, IN, USA}

\cortext[corxh]{Corresponding author. Tel: +1 540 3156512}
\ead{jwang33@nd.edu}

\begin{abstract}
The hybrid neural differentiable models mark a significant advancement in the field of scientific machine learning. These models, integrating numerical representations of known physics into deep neural networks, offer enhanced predictive capabilities and show great potential for data-driven modeling of complex physical systems. However, a critical and yet unaddressed challenge lies in the quantification of inherent uncertainties stemming from multiple sources. Addressing this gap, we introduce a novel method, DiffHybrid-UQ, for effective and efficient uncertainty propagation and estimation in hybrid neural differentiable models, leveraging the strengths of deep ensemble Bayesian learning and nonlinear transformations. Specifically, our approach effectively discerns and quantifies both aleatoric uncertainties, arising from data noise, and epistemic uncertainties, resulting from model-form discrepancies and data sparsity. This is achieved within a Bayesian model averaging framework, where aleatoric uncertainties are modeled through hybrid neural models. The unscented transformation plays a pivotal role in enabling the flow of these uncertainties through the nonlinear functions within the hybrid model. In contrast, epistemic uncertainties are estimated using an ensemble of stochastic gradient descent (SGD) trajectories. This approach offers a practical approximation to the posterior distribution of both the network parameters and the physical parameters. Notably, the DiffHybrid-UQ framework is designed for simplicity in implementation and high scalability, making it suitable for parallel computing environments. The merits of the proposed method have been demonstrated through problems governed by both ordinary and partial differentiable equations.

\end{abstract}

\end{frontmatter}



\section{Introduction}
\label{sec:intro}


The evolving dynamics of computational science are driven by the advent of advanced numerical algorithms, enhanced computational infrastructures, and the proliferation of extensive datasets. This evolution has catalyzed the development of innovative methodologies to address multifaceted challenges associated with scientific modeling and predictive analytics. Within this transformative framework, Scientific Machine Learning (SciML) emerges as a salient discipline, intricately blending the principles of traditional scientific modeling with the state-of-the-art machine learning techniques, especially in the context of handling voluminous data and advanced GPU computing. At its core, SciML represents a paradigm that seamlessly integrates the rigorous foundations of scientific principles with the adaptability  and efficiency of machine learning methods. Such an integration not only amplifies the computational efficiency but often enhances the accuracy of predictions for complex scientific problems. Illustrative of the advancements within the SciML domain are methodologies like Physics-Informed Neural Networks (PINN)~\cite{raissi2019physics,sun2020surrogate,sun2020physics}, neural operators~\cite{li2021fourier,lu2021learning,wang2021learning}, equation discovery techniques~\cite{brunton2016discovering,chen2021physics,sun2022bayesian}, and hybrid neural models~\cite{kochkov2021machine,liu2022predicting,fan2023differentiable,akhare2023physics}.

In the expanding landscape of SciML, hybrid neural modeling emerges as a distinctive frontier that marries domain-specific knowledge with cutting-edge data-driven methodologies, striking a balance between deep-rooted scientific principles and the adaptability of modern machine learning (ML) techniques. While traditional scientific models, rooted in established physical laws and equations, are invaluable tools for understanding various physical phenomena, they often struggle when faced with complex systems whose underlying physics has not been fully understood or when managing extensive datasets. In contrast, deep neural networks (DNNs) are adept at deciphering intricate patterns from large-scale data, but their black-box nature can pose interpretability challenges and sometimes limits their generalization in out-of-sample scenarios. Combining the strengths of traditional scientific models with ML, hybrid techniques endeavor to address these challenges, leveraging the advantages of both worlds. Past effort in hybrid learning model tend to weakly integrate ML with physics-based numerical models~\cite{tompson2017accelerating,vinuesa2022enhancing}. Typically, an ML model is trained offline, and later incorporated into a standard numerical solver. Such strategies gained traction, especially in data-driven closure modeling for turbulence~\cite{wang2017physics,duraisamy2019turbulence,wang2019prediction} or atmospheric flows~\cite{zanna2020data,mendez2023data}. A significant challenge, however, arises from the necessity for labeled data during intermediate phases –- a requirement that is frequently unmet. Furthermore, when ML models and numerical solvers operate somewhat independently, the resulting hybrid system can sometimes exhibit instability, leading to unreliable and often inaccurate \emph{a posteriori} predictions~\cite{wu2019reynolds}. 

Recognizing these challenges, the prevailing consensus highlights the advantages of a more integrated approach -- creating a unified hybrid model that promises not just to improve computational efficiency, but also yield more robust and accurate solutions. The paradigm of differentiable programming (DP) offers a pathway to this goal, enabling joint optimization of both ML and numerical components within a unified training environment. Recently, there has been a growing interest in developing differentiable solvers and hybrid neural models, which have demonstrated significant potential across various scientific fields~\cite{innes2019differentiable,belbute2020combining,huang2020learning,kochkov2021machine,list2022learned,akhare2023physics,liu2022predicting,fan2023differentiable}. For example, Kochkov et al.~\cite{kochkov2021machine} integrated convolutional neural networks into a differentiable computational fluid dynamic (CFD) solver, facilitating more effective coarse graining. Similarly, Huang et al.~\cite{huang2020learning} incorporated DNNs into a differentiable finite element solver, enabling the derivation of constitutive relations for nonlinear materials from indirect measurements. Wang and co-workers have pioneered differentiable hybrid neural models to efficiently emulate spatiotemporal physics of fluid~\cite{liu2022predicting}, fluid-structure interaction~\cite{fan2023differentiable}, and composite materials manufacturing processes~\cite{akhare2023physics}. Their designs are deeply rooted in the profound relationship between neural network architectural components (e.g., convolution layers, residual connections) and the numerical representations of PDEs - a subject currently gaining traction in the ML community~\cite{chen2018neural,long2018pde,zhang2020forward,eliasof2020diffgcn,thorpe2021grand++}. Building on these connections, they pointed out that traditional numerical solvers can be interpreted as unique neural network instances, where the convolutional kernels, recurrent structures, and residual connections are pre-determined entirely by the known physics and associated numerical schemes, rather than being learned from data~\cite{liu2022predicting}. Therefore, the differentiable hybrid neural models can be viewed as physics-integrated neural networks, where the known physics, represented by discretized PDE operators, are woven into the neural architecture, guided by the connection between numerical PDEs and neural architecture components~\cite{fan2023differentiable}.

Despite the promise of hybrid neural modeling in advancing computational science and engineering, addressing the intrinsic uncertainties in these models remains a significant concern. These models, while effectively integrating domain-specific knowledge with machine learning methodologies, are not immune to uncertainties from multiple sources. For example, aleatoric uncertainty can emerge from noisy training datasets, while the vast parameter landscape of neural networks and potential model-form discrepancy of physics-based model can induce epistemic uncertainty. The growing prominence of these issues has brought uncertainty quantification (UQ) in SciML into the spotlight~\cite{psaros2023uncertainty}. As we know, UQ for DNNs is a long-standing challenge within the ML community, largely due to the intricate, nonlinear nature of DL models. The extremely high-dimensional parameter space of DL models make traditional Bayesian sampling techniques infeasible~\cite{abdar2021review}. This limitation has directed researchers toward approximate Bayesian learning strategies such as variational inference~\cite{blei2017variational}, stochastic gradient MCMC~\cite{welling2011bayesian,zhang2020amagold}, and Bayes by back-propagation~\cite{blundell2015weight}. In parallel, ensemble learning methods such as deep ensemble~\cite{lakshminarayanan2017simple}, snapshot ensemble~\cite{huang2017snapshot}, SWAG~\cite{maddox2019simple}, and SeBayS~\cite{jantre2022sequential} have emerged as effective tools for estimating uncertainty in DNNs. The posterior landscape, however, becomes even more intricate for hybrid neural models that blend physics-based numerical solvers with DNNs. The fusion of mathematical representations of physical phenomena with the inherent complexities of neural networks introduces multifaceted sources of uncertainty. How to quantify and propagate uncertainty for these differentiable hybrid neural models remains uncharted territory, indicating a significant gap in the current literature.  

In this study, we present DiffHybrid-UQ, a novel probabilistic differentiable hybrid neural modeling framework designed for predicting spatiotemporal physics. This framework stands out due to its capability of quantifying associated uncertainty propagation in a scalable manner. At the core of our proposed method is the integration of the deep-ensemble based stochastic weight averaging Gaussian (SWAG) method with the unscented nonlinear transformation, devised to effectively estimate and propagate multi-source uncertainties inherent to hybrid neural models, offering both robust predictions and comprehensive uncertainty assessments. Specifically, our method discerns both aleatoric and epistemic uncertainties stemming from diverse sources, such as measurement noise and model form discrepancies, within a Bayesian model averaging framework. Aleatoric uncertainties, grounded on the Gaussian assumption, are modeled using DNNs, where the associated hyperparameters are learned from data. The unscented transformation is adopted to facilitate the flow of uncertainty through the nonlinear functions within the hybrid framework. On the other hand, epistemic uncertainties are estimated using an ensemble of stochastic gradient decent (SGD) trajectories, providing an approximation to the posterior distribution of the network and physical parameters. A salient advantage of the proposed DiffHybrid-UQ is its straightforward implementation and amenability to parallelism. Moreover, the design sidesteps extensive hyperparameter tuning, minimizing chances of biased outcomes. This endeavor aims to fill the research gap concerning uncertainty propagation and quantification in physics-integrated differentiable hybrid neural models, bolstering the reliability and robustness of their predictions. The rest of the paper is organized as follows: Section~\ref{sec:methodology} detailed the proposed methodology, including problem formulation from a Bayesian perspective and modeling of both aleatoric and epistemic uncertainties. The merits of the proposed model have been demonstrated in a variety of numerical experiments in Section~\ref{sec:result}
Finally, Section~\ref{sec:conclusion} concludes the paper.

\section{Methodology}
\label{sec:methodology}

\subsection{Problem formulation}
\label{sec:problem}

In various scientific and engineering disciplines, Partial Differential Equations (PDEs) are ubiquitously employed as a mathematical formalism to describe the behavior of continua, including but not limited to fluid dynamics, solid mechanics, heat and mass transfer, electromagnetic fields, and quantum mechanical systems. However, for many complex systems, the analytical form of the governing PDEs is often partially known, which can be generally represented as,
\begin{linenomath*}
\begin{subequations}
    \begin{alignat}{3}
    \frac{\partial \mathbf{v}}{\partial t} = \mathscr{F}\big[\mathcal{K} (\mathbf{v}(\bar{\mathbf{x}}); \boldsymbol{\lambda}_{\mathcal{K}}, \boldsymbol{\lambda}_{\mathcal{U}}), \mathcal{U} (\mathbf{v}(\bar{\mathbf{x}}); \boldsymbol{\lambda}_{\mathcal{K}}, \boldsymbol{\lambda}_{\mathcal{U}})\big], \ \ \ \ &\bar{\mathbf{x}} \in  \Omega_{p,t}
    \label{eq:PiNDiff-PDE},\\
    \mathcal{BC}(\mathbf{v}(\bar{\mathbf{x}}); \boldsymbol{\lambda}_{\mathcal{K}}, \boldsymbol{\lambda}_{\mathcal{U}}) = 0,\ \  \ \ &\bar{\mathbf{x}} \in  \partial\Omega_{p,t}
    \label{eq:PiNDiff-BC},\\
    \mathcal{IC}(\mathbf{v}(\bar{\mathbf{x}}); \boldsymbol{\lambda}_{\mathcal{K}}, \boldsymbol{\lambda}_{\mathcal{U}}) = 0,\ \  \ \ &\bar{\mathbf{x}} \in  \Omega_{p, t=0}
    \label{eq:PiNDiff-IC},
    \end{alignat}
    \label{eq:PiNDiff}
\end{subequations}
\end{linenomath*}
where $\bar{\mathbf{x}} = \{\textbf{x}, t\}$ represents the spatiotemporal coordinates, with $\mathbf{x}$ specifically denoting spatial coordinates and $t$ representing time. The spatiotemporal domain $\Omega_{p,t}$ is defined as the Cartesian product $\Omega_{p} \times[0, T_r]$, where $\Omega_{p}$ represents the spatial extent of the physical domain and $T_r$ denotes the time range under consideration; $\partial\Omega_{p,t} \triangleq \partial\Omega_{p} \times[0, T_r]$ with $\partial\Omega_{p}$ representing the boundary of the physical domain. The nonlinear functions $\mathcal{K}(\cdot)$ and $\mathcal{U}(\cdot)$ characterize the known and unknown segments of the PDEs, respectively, which are coupled by a nonlinear functional $\mathscr{F}(\cdot)$. These functions are intrinsically reliant on certain physical parameters, which may sometimes remain elusive or uncertain. In this formulation, $\boldsymbol{\lambda}_\mathcal{K}$ and $\boldsymbol{\lambda}_\mathcal{U}$ denote the known and unknown physical parameters, respectively. Furthermore, the initial and boundary conditions can be abstractly represented by the differential operators $\mathcal{IC}$ and $\mathcal{BC}$, respectively, which may also be contingent upon specific parameters.

Direct physics-based modeling faces substantial challenges due to the incompleteness of the governing equation. Addressing this intricate issue, differential hybrid (DiffHybrid) models such as PiNDiff model~\cite{akhare2023physics} have emerged to integrate neural networks with the known part of the PDEs using differentiable programming (DP) to learn the physics from data. Within this paradigm, unknown functions/operators such as $\mathscr{F}(\cdot)$ and $\mathcal{U}(\cdot)$, are modeled as DNNs -- $\mathscr{F}_{nn}$ and $\mathcal{U}_{nn}$ -- with trainable parameters $\boldsymbol{\theta}_{nn}$. Simultaneously, the elusive physical parameters $\boldsymbol{\lambda}_{\mathcal{U}}$ are rendered trainable. A salient feature of the DiffHybrid models lies in their construction using DP, allowing the joint optimization of both DNNs and unknown physical parameters $\boldsymbol{\lambda}_{\mathcal{U}}$ within a unified training environment through gradient back-propagation. The training of DiffHybrid model can be formulated as a PDE-constrained optimization problem, where the parameters to be optimized are collectively represented as $\boldsymbol{\theta} = [\boldsymbol{\theta}_{nn}, \boldsymbol{\lambda}_{\mathcal{U}}]^T \in \mathbb{R}^d$. Given data $\mathcal{D}=\{\bar{\mathbf{x}}_i,\mathbf{v}_i\}^N_{i=1}$, these trainable parameters are optimized by minimizing the objective function, $J(\mathbf{v}_{\boldsymbol{\theta}}(\bar{\mathbf{x}}), \mathcal{D})$, defined, for example, by MSE or negative log-likelihood. The optimization procedure is expressed as follows,
\begin{equation}
\begin{aligned}
\min_{\boldsymbol{\theta}} \quad & J(\mathbf{v}_{\boldsymbol{\theta}}(\bar{\mathbf{x}}), \mathcal{D}; \boldsymbol{\theta})\\
\textrm{s.t.} \quad & 
    \frac{\partial \mathbf{v}_{\boldsymbol{\theta}}(\bar{\mathbf{x}})}{\partial t} = \mathscr{F}_{nn}\big[\mathcal{K}(\mathbf{v}_{\boldsymbol{\theta}}(\bar{\mathbf{x}}); \boldsymbol{\lambda}_{\mathcal{K}}, \boldsymbol{\lambda}_{\mathcal{U}}), \mathcal{U}_{nn}(\mathbf{v}_{\boldsymbol{\theta}}(\bar{\mathbf{x}}); \boldsymbol{\lambda}_{\mathcal{K}}, \boldsymbol{\lambda}_{\mathcal{U}},\boldsymbol{\theta}_{nn}); \boldsymbol{\theta}_{nn}\big], \ \ \ \ \bar{\mathbf{x}} \in  \Omega_{p,t} \\
  &
    \mathcal{BC}(\mathbf{v}_{\boldsymbol{\theta}}(\bar{\mathbf{x}}); \boldsymbol{\lambda}_{\mathcal{K}}, \boldsymbol{\lambda}_{\mathcal{U}}) = 0,\ \  \ \ \bar{\mathbf{x}} \in  \partial\Omega_{p,t}\\
  &
    \mathcal{IC}(\mathbf{v}_{\boldsymbol{\theta}}(\bar{\mathbf{x}}); \boldsymbol{\lambda}_{\mathcal{K}}, \boldsymbol{\lambda}_{\mathcal{U}}) = 0,\ \  \ \ \bar{\mathbf{x}} \in  \Omega_{p, t=0}
\end{aligned}
\label{eq:Hybrid_opt}
\end{equation}
As a result of the end-to-end training process, the DiffHybrid model ensures that the solution vector $\mathbf{v}_{\boldsymbol{\theta}}(\bar{\mathbf{x}}) = \mathrm{DiffHybrid}(\bar{\mathbf{x}};\boldsymbol{\theta})$ consistently satisfies the governing PDEs for the given set of parameters $\boldsymbol{\theta}$. While DiffHybrid models, as described above, exhibit significant advantages over conventional black-box DNNs in predictive performance~\cite{akhare2023physics}, they are not without vulnerabilities, especially when applied to out-of-training scenarios. This limitation raises questions about the reliability and robustness of the model's predictions, making the estimation and propagation of uncertainty through hybrid neural solvers a crucial yet challenging endeavor. To this end, we propose the \emph{DiffHybrid-UQ} framework, which employs an ensemble-based Stochastic Weight Averaging Gaussian (En-SWAG) method to estimate the epistemic uncertainty and utilizes the unscented transformation technique for the propagation of Gaussian-based aleatoric uncertainty through the nonlinear functions. This approach aims to substantially improve the reliability and robustness of DiffHybrid models by offering not just accurate but also uncertainty-quantified predictions.

\subsection{Overview of DiffHybrid-UQ framework}
To effectively quantify the inherent uncertainties associated with the predicted state variable $\mathbf{v}_{\boldsymbol{\theta}}(\tilde{\mathbf{x}})$, it is modeled as a random variable, denoted by $\mathbf{V}(\tilde{\mathbf{x}})$ conditioned on modeling parameters $\boldsymbol{\theta}$, where $\tilde{\mathbf{x}}$ represents the inputs of the model including spatiotemporal coordinates $\bar{\mathbf{x}}$ and other nontrainable input parameters in the governing PDEs and IC/BCs. In this framework, the hybrid neural model is designed to yield the probability distribution $p(\mathbf{V} | \tilde{\mathbf{x}}, \boldsymbol{\theta})$ for random variable $\mathbf{V}$ corresponding to a specified input $\tilde{\mathbf{x}}$ and associated model parameters $\boldsymbol{\theta}$. Given a dataset $\mathcal{D}$, the model can be trained following Bayesian principles, which is to determine the posterior distribution of the modeling parameters as $p(\boldsymbol{\theta}|\mathcal{D})$. Namely, the output of the trained DiffHybrid-UQ model is the probability distribution $p(\mathbf{V}|\tilde{\mathbf{x}}, \mathcal{D})$ of the predicted states $\mathbf{V}$, conditional on the training dataset $\mathcal{D}$, which can be derived using Bayesian Model Averaging (BMA),
\begin{equation}
    p(\mathbf{V}|\tilde{\mathbf{x}} , \mathcal{D}) = \int p(\mathbf{V} | \tilde{\mathbf{x}}, \boldsymbol{\theta}) p(\boldsymbol{\theta} | \mathcal{D}) d\boldsymbol{\theta},
    \label{eq:pred}
\end{equation}
where posterior distribution $p(\boldsymbol{\theta} | \mathcal{D})$ of trainable parameters is computed using Bayes' theorem, 
\begin{equation}
    p(\boldsymbol{\theta}|\mathcal{D}) = \frac{p(\mathcal{D|\boldsymbol{\theta}})p(\boldsymbol{\theta})}{\int p(\mathcal{D}|\boldsymbol{\theta}) p(\boldsymbol{\theta}) d\boldsymbol{\theta}},
    \label{eq:posterior}
\end{equation}  
where $p(\boldsymbol{\theta})$ is our prior beliefs about $\boldsymbol{\theta}$ and $p(\mathcal{D|\boldsymbol{\theta}})$ is the joint likelihood of the datasest given the model prediction $\mathbf{V}$ with specific $\tilde{\mathbf{x}}$ and $\boldsymbol{\theta}$. As the measurement noise of the dataset $\mathcal{D}=\{\tilde{\mathbf{x}}_i,\mathbf{v}_{\mathcal{D},i}\}^N_{i=1}$ is usually independent and identically distributed, the joint likelihood can be expressed as $p(\mathcal{D|\boldsymbol{\theta}}) = \prod_{i=1}^N p(\mathbf{v}_{\mathcal{D},i}|\tilde{\mathbf{x}}_i, \boldsymbol{\theta})$.

The integral in Eq.~\ref{eq:pred} can be estimated via Monte Carlo method in practice,
\begin{equation}
    p(\mathbf{V}|\tilde{\mathbf{x}} , \mathcal{D}) \approx \frac{1}{M}\sum_{j=1}^M p(\mathbf{V} | \tilde{\mathbf{x}}, \boldsymbol{\theta}^{(j)}), \quad \boldsymbol{{\theta}}^{(j)} \sim p(\boldsymbol{\theta} | \mathcal{D}),
    \label{eq:bma}
\end{equation}
where $\{\boldsymbol{\theta}^{(j)}\}_{j=1}^M$ are $M$ Monte Carlo samples obtained from the posterior distribution $p(\boldsymbol{\theta} | \mathcal{D})$. Through this approximation, the probabilistic prediction of $(\mathbf{V}|\tilde{\mathbf{x}} , \mathcal{D})$ essentially becomes a mixture of density functions conditioned on $\boldsymbol{\theta}$ samples. Typically, one of the primary metrics to express uncertainty is the variance (often used to define the credible interval), which can be decomposed as follows using the law of total variance:
\begin{equation}
    \mathbb{V}\mathrm{ar}(\mathbf{V}|\tilde{\mathbf{x}} , \mathcal{D}) = \underbrace{\mathbb{E}_{\boldsymbol{\theta}|\mathcal{D}}[\mathbb{V}\mathrm{ar}(\mathbf{V}|\tilde{\mathbf{x}}, \boldsymbol{\theta})]}_{\text{Aleatoric}} + \underbrace{\mathbb{V}\mathrm{ar}_{\boldsymbol{\theta}|\mathcal{D}}(\mathbb{E}[\mathbf{V}|\tilde{\mathbf{x}}, \boldsymbol{\theta}])}_{\text{Epistemic}},
    \label{eq:uqsplit}
\end{equation}
where the aleatoric uncertainty captures inherent system variabilities like measurement noise, while the epistemic uncertainty reflects the inadequacy of our model, stemming from incomplete physical understanding and the intricacies of neural network predictions. Up to now, two primary challenges still remain unresolved: (1) effectively modeling the likelihood function to capture the aleatoric uncertainty; and (2) addressing the practical difficulties in sampling the posterior $p(\boldsymbol{\theta}|\mathcal{D})$, given the extremely high dimensionality of the parameter space inherent to the hybrid neural model. Subsequent subsections will delve into our solutions for these challenges within the proposed DiffHybrid-UQ framework.

\subsection{Modeling likelihood via DiffHybrid model for aleatoric UQ}
To account for aleatoric uncertainty, the measurement noise $\boldsymbol{\epsilon}$ of the state variables is assumed to be an additive Gaussian noise with zero mean,
\begin{equation}
    \mathbf{V}|\tilde{\mathbf{x}}, \boldsymbol{\theta} = \bar{\mathbf{v}}_{\boldsymbol{\theta}}(\tilde{\mathbf{x}}) + \boldsymbol{\epsilon}_{\boldsymbol{\theta}}.
    \label{eq:measurement-model}
\end{equation}
Namely, the conditional probability of the state $(\mathbf{V}|\tilde{\mathbf{x}}, \boldsymbol{\theta})$ is Gaussian,
\begin{equation}
    p(\mathbf{V}|\tilde{\mathbf{x}}, \boldsymbol{\theta}) = \mathcal{N}\bigg(\bar{\mathbf{v}}_{\boldsymbol{\theta}}(\tilde{\mathbf{x}}), {diag}(\Sigma_{\mathbf{V},\boldsymbol{\theta}})(\tilde{\mathbf{x}})\bigg),
    \label{eq:measurement-model2}
\end{equation}
where the mean $\bar{\mathbf{v}}_{\boldsymbol{\theta}}(\tilde{\mathbf{x}})$ and diagonal covariance matrix ${diag}(\Sigma_{\mathbf{V},\boldsymbol{\theta}})(\tilde{\mathbf{x}})$ are approximated by the differentiable hybrid neural model with trainable parameters $\boldsymbol{\theta}$. Alternatively put, the output of the DiffHybrid solver is the probability distribution of the predicted state variable, characterized by it's first and second moments,
\begin{equation}
    \bar{\mathbf{v}}_{\boldsymbol{\theta}}(\tilde{\mathbf{x}}), {diag}(\Sigma_{\mathbf{V},\boldsymbol{\theta}})(\tilde{\mathbf{x}}) = \mathrm{DiffHybrid}(\tilde{\mathbf{x}};\boldsymbol{\theta}).
    \label{eq:diffHybridUQ}
\end{equation}
Given the dataset $\mathcal{D}=\{\tilde{\mathbf{x}}_i,\mathbf{v}_{\mathcal{D},i}\}^N_{i=1}$, the joint likelihood is computed as,
\begin{equation}
    p(\mathcal{D|\boldsymbol{\theta}}) = \prod_{i=1}^N \frac{1}{\sqrt{(2\pi)^d \det{(\Sigma_{\mathbf{V},\boldsymbol{\theta}})}}} \exp{\bigg(-\frac{1}{2} \big(\mathbf{v}_{\mathcal{D},i} -\bar{\mathbf{v}}_{\boldsymbol{\theta}}(\tilde{\mathbf{x}}_i)\big)^T \Sigma_{\mathbf{V},\boldsymbol{\theta}}^{-1}(\tilde{\mathbf{x}}_i) \big(\mathbf{v}_{\mathcal{D},i} -\bar{\mathbf{v}}_{\boldsymbol{\theta}}(\tilde{\mathbf{x}}_i)\big) \bigg)},
    \label{eq:likelihood-gaussian}
\end{equation}
where $d$ is the dimension of the spatiotemporal state. When disregarding the uncertainty associated with $\boldsymbol{\theta}$, the training of the DiffHybrid model can be formulated as the minimization of the negative log-likelihood,
\begin{equation}
\boldsymbol{\theta}^* =  \min_{\boldsymbol{\theta}} \quad  J\big(\mathcal{D}, (\mathbf{V} | \tilde{\mathbf{x}}, \boldsymbol{\theta})\big) = \min_{\boldsymbol{\theta}} \quad -\log{p(\mathcal{D}|\boldsymbol{\theta})}.
\end{equation}
This process allows the model to quantify the uncertainty arising from the noise present in the dataset $\mathcal{D}$. However, a crucial point to bear in mind is that this training merely yields a single realization, i.e., a point estimate $\boldsymbol{\theta}^*$, out of the full posterior distribution $p(\boldsymbol{\theta} | \mathcal{D})$. Consequently, this model is unable to capture epistemic uncertainty that originates from DiffHybrid model parameterization. To truly gauge this, it becomes imperative to also approximate the parameter posterior and assess the total uncertainty as described in Eq.~\ref{eq:bma}. Using Monte Carlo based BMA, the probability distribution of the prediction $p(\mathbf{V}|\tilde{\mathbf{x}}, \mathcal{D})$ given data becomes a Gaussian mixture, with its mean and variance (i.e., diagonal part of its covariance matrix) as, 
\begin{subequations}
    \begin{alignat}{2}
    \mathbb{E}(\mathbf{V}|\tilde{\mathbf{x}}, \mathcal{D}) &\approx \frac{1}{M} \sum^M_{j=1} \bar{\mathbf{v}}_{\theta}(\tilde{\mathbf{x}};\boldsymbol{{\theta}}^{(j)}),\\
    \mathbb{V}\mathrm{ar}(\mathbf{V}|\tilde{\mathbf{x}}, \mathcal{D}) &\approx \underbrace{\frac{1}{M} \sum^M_{j=1} diag \big(\Sigma_{\mathbf{V},\boldsymbol{\theta}}(\tilde{\mathbf{x}};\boldsymbol{{\theta}}^{(j)})\big)}_{\text{Aleatoric}} + \underbrace{\frac{1}{M} \sum^M_{j=1} \big(\bar{\mathbf{v}}_{\theta}(\tilde{\mathbf{x}};\boldsymbol{{\theta}}^{(j)}) - \mathbb{E}(\mathbf{V}|\tilde{\mathbf{x}}, \mathcal{D})\big)^2}_{\text{Epistemic}}
    \end{alignat}
    \label{eq:pred_approx}
\end{subequations}
Section~\ref{sec:epistemicUQ} will delve into the practical methods for approximating the posterior $p(\boldsymbol{\theta} | \mathcal{D})$.

\subsection{Aleatoric uncertainty propagation within DiffHybrid framework}
\label{sec:UP}
\begin{figure}[!ht]
\centering
{\includegraphics[width=1.\textwidth]{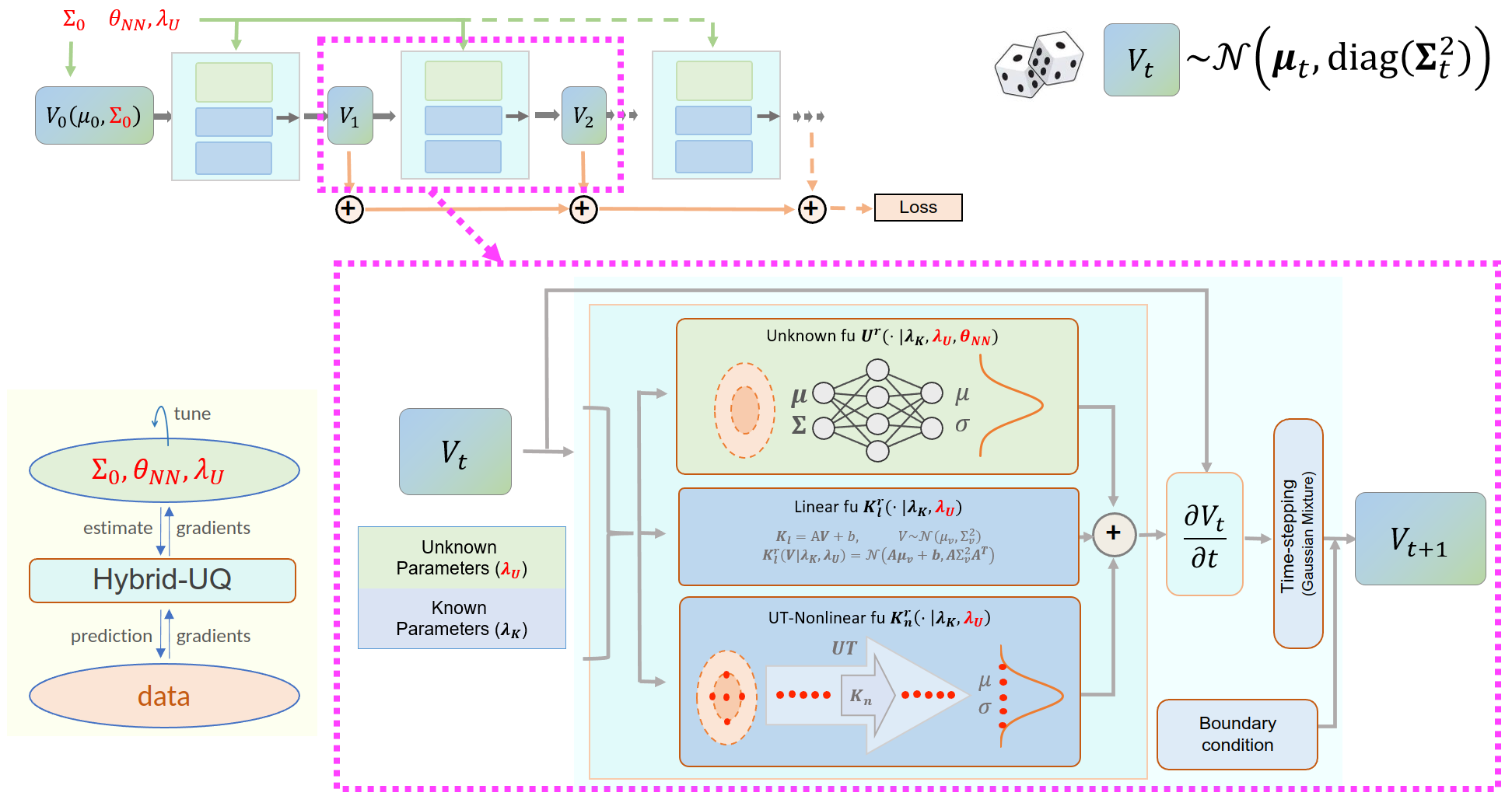}}
\caption{The overview of the auto-regressive DiffHybrid architecture for uncertainty propagation.}
\label{fig:DiffHybrid-UQ}
\end{figure}

In this subsection, we elaborate on the design of the DiffHybrid framework for likelihood modeling. As indicated in Eq.~\ref{eq:diffHybridUQ}, the output of the DiffHybrid model represents the probability distribution of the predicted state variable, which is primarily characterized by its mean and variance. This necessitates the formulation of a model adept at processing and predicting these statistical attributes (i.e., first- and second-order moments). As discussed in Section~\ref{sec:problem}, the DiffHybrid framework is a implicit neural network defined by the partially known governing PDEs. Namely, the stochastic state $\mathbf{V}_{\theta}(\bar{\mathbf{x}})$ is modeled implicitly as,
\begin{linenomath*}
\begin{subequations}
    \begin{alignat}{3}
    \frac{\partial \mathbf{V}_{\boldsymbol{\theta}}(\bar{\mathbf{x}})}{\partial t} = \mathscr{F}_{nn}\big[\mathcal{K}(\mathbf{V}_{\boldsymbol{\theta}}(\bar{\mathbf{x}}); \boldsymbol{\lambda}_{\mathcal{K}}, \boldsymbol{\lambda}_{\mathcal{U}}), \mathcal{U}_{nn}(\mathbf{V}_{\boldsymbol{\theta}}(\bar{\mathbf{x}}); \boldsymbol{\lambda}_{\mathcal{K}}, \boldsymbol{\lambda}_{\mathcal{U}},\boldsymbol{\theta}_{nn}); \boldsymbol{\theta}_{nn}\big], \ \ \ \ &\bar{\mathbf{x}} \in  \Omega_{p,t}
    \label{eq:DiffHybrid-UQ-PDE},\\
    \mathcal{BC}(\mathbf{V}_{\boldsymbol{\theta}}(\bar{\mathbf{x}}); \boldsymbol{\lambda}_{\mathcal{K}}, \boldsymbol{\lambda}_{\mathcal{U}}) = 0,\ \  \ \ &\bar{\mathbf{x}} \in  \partial\Omega_{p,t}
    \label{eq:DiffHybrid-UQ-BC},\\
    \mathcal{IC}(\mathbf{V}_{\boldsymbol{\theta}}(\bar{\mathbf{x}}); \boldsymbol{\lambda}_{\mathcal{K}}, \boldsymbol{\lambda}_{\mathcal{U}}, \Sigma^2_0) = 0,\ \  \ \ &\bar{\mathbf{x}} \in  \Omega_{p, t=0}
    \label{eq:DiffHybrid-UQ-IC},
    \end{alignat}
    \label{eq:DiffHybrid-UQ}
\end{subequations}
\end{linenomath*}
When formulating auto-regressive structure for temporal evolution, the architecture of the DiffHybrid neural for uncertainty propagation is depicted in Fig.~\ref{fig:DiffHybrid-UQ}.
In contrast to deterministic DiffHybrid models~\cite{akhare2023physics,fan2023differentiable}, the operators $\mathcal{K}$, $\mathcal{U}_{nn}$, and $\mathscr{F}_{nn}$ here are designed to handle probabilistic distribution rather than deterministic values. To this end, all neural networks within this architecture are configured to process the mean and covariance at each grid point as inputs, producing dual outputs that represent the estimated mean and covariance of the intermediate functions. This design enables the DiffHybrid model to learn the spatiotemporal dynamics via the prediction mean $\bar{\mathbf{v}}_{\theta}(\bar{\mathbf{x}})$, and also to quantify the associated aleatoric uncertainty via the diagonal covariance estimation, $\text{diag}\big(\Sigma_{V,\theta}^2(\bar{\mathbf{x}})\big)$, which are spatiotemporal fields as well. The initial diagnoal covariance values are set as $\text{diag}\big(\Sigma_{V,\theta}^2({\mathbf{x}}, t=0)\big) = \boldsymbol{\theta}_{\Sigma^2_0}$, which are trainable. These values will be propagated over time following the autoregressive temporal evolution, as illustrated in Fig.~\ref{fig:DiffHybrid-UQ}. In the current model design, the $\text{diag}(\Sigma^2)$ value at the boundary is assumed to be zero. The trainable parameters of the DiffHybrid-UQ model include DNN parameters, unknown physical coefficient, initial diagonal covariance matrix, collectively denoted as $\boldsymbol{\theta} = (\boldsymbol{\theta}_{\lambda_{U}}, \boldsymbol{\theta}_{\mathcal{NN}}, \boldsymbol{\theta}_{\Sigma^2_0})$. Detailed DNNs architecture used for this study can be found in \ref{sec:NNfu}.

The PDE operators within the known segment of the right-hand side of the governing equation (Eq.\ref{eq:DiffHybrid-UQ}) require discretization to formulate the predefined convolution layers, encoding the known physics. Detailed information on the discretization process is available in \ref{sec:Discretization}. After discretization, these discretized known operators need to manage the propagation of uncertainty on the grid level. We consider two types of operators in this context: linear and nonlinear known PDE operators. For linear operators $\mathcal{K}_l$, such as discretized diffusion terms $\nabla^2$, the propagation of mean and covariance matrix of a random variable $V$ is straightforward. This is because the affine transformation of a Gaussian distribution result in another Gaussian distribution. Specifically, we have
\begin{equation}
 \mathcal{K}_l \big(\mathcal{N}(\bar{\mathbf{v}}_{\theta}, \Sigma_{V,\theta}^2)\big) = \mathcal{N}(K \bar{\mathbf{v}}_{\theta} + b , K \Sigma_{V,\theta}^2 K^T),
\end{equation}
where the linear PDE operator is defined as $\mathcal{K}_l(\mathbf{v}) = K\mathbf{v}+b$. 

\begin{figure}[!ht]
\centering
{\includegraphics[width=0.5\textwidth]{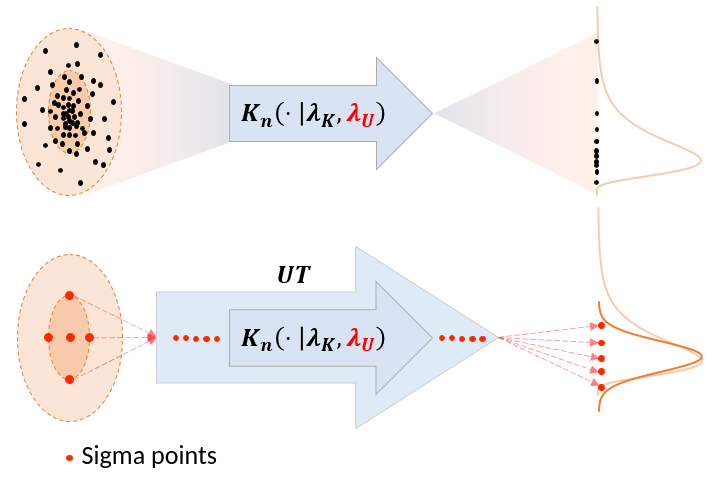}}
\caption{Illustrative plot of unscented transformation (UT) versus Monte Carlo (MC) methods for uncertainty propagation.}
\label{fig:UT}
\end{figure}

When it comes to noninear PDE operators, $\mathcal{K}_n$, the propagation of mean and variance is more challenging, as it typically requires either linearization of the operator, which often compromises accuracy, or the use of Monte Carlo sampling methods, which can be computationally intensive if not infeasible. To address these challenges, our work utilizes the unscented transform method~\cite{wan2000unscented}. The unscented transform (UT) is a mathematical technique used for filtering and smoothing in nonlinear dynamical systems. It aims to accurately propgate the first and second moments of a probability distribution through nonlinear mappings. The method involves generating a small, deterministically selected set of sigma points that effectively capture the mean and covariance of the initial probability distribution. These sigma points are then transformed by applying the target nonlinear function to each point, a process that distinguishes the UT from Monte Carlo methods which rely on random sampling. The number of sigma points used is small, greatly enhancing efficiency over Monte Carlo methods. After undergoing the nonlinear transformation, the mean and covariance of the new distribution are recalculated from these transformed sigma points through a weighted summation, with weights as specified in~\cite{wan2000unscented}. This method is especially advantageous as it avoids the complexities of Jacobian matrix calculations required in linearization techniques like the Extended Kalman Filter. Moreover, it yields higher accuracy in scenarios with significant nonlinearities or when precision in the distribution's tail is crucial. A illustrative plot of the UT vs. MC based uncertainty propagation is shown in Fig.~\ref{fig:UT}. More implementation details can be found in \ref{sec:UTfun}.

\subsection{Scalable posterior sampling for epistemic UQ}
\label{sec:epistemicUQ}

\begin{figure}[!ht]
    \centering
    \includegraphics[width = 1.0\textwidth]{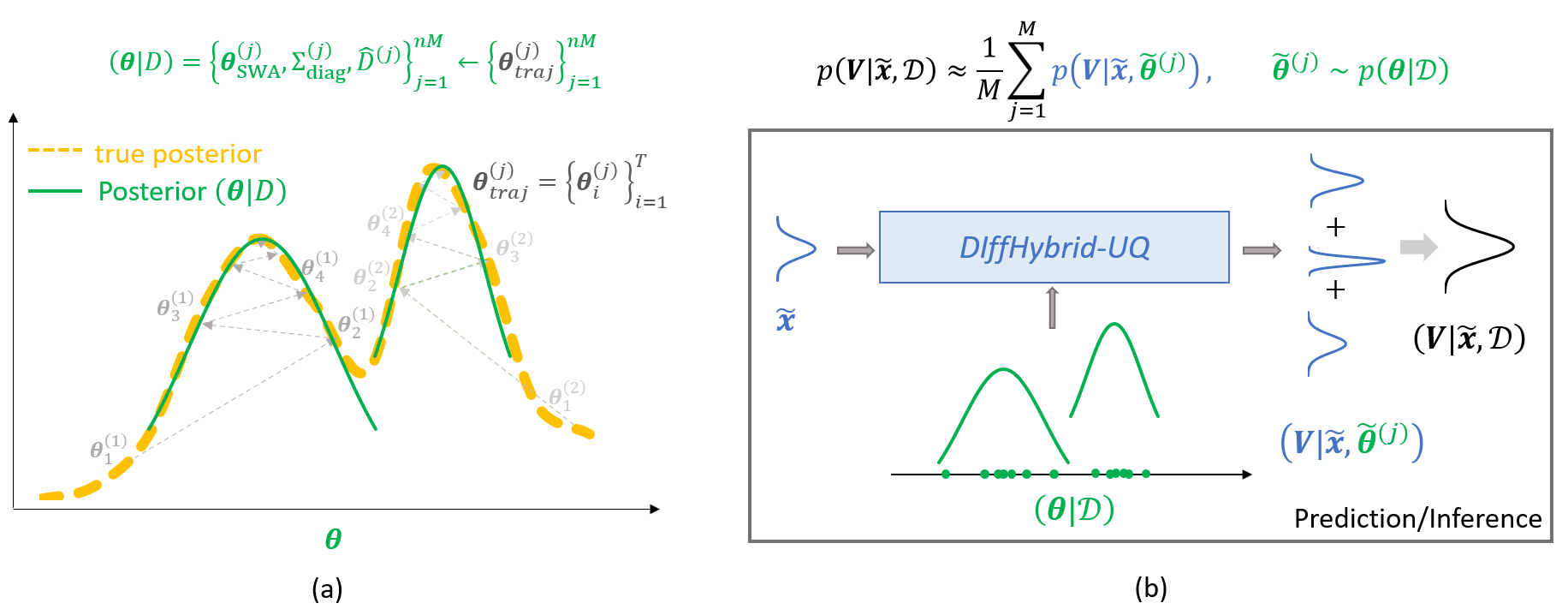}
    \caption{(a) A schematic of the estimation of the posterior distribution using the ensemble-based SWAG training (b) Prediction/Inference using the posterior distribution}
    \label{fig:SWAG_pred}
\end{figure}

Direct computation of the posterior $p(\boldsymbol{\theta}|\mathcal{D})$ using traditional Bayesian sampling techniques, e.g., MCMC, is impractical for deep learning models due to their high dimensionality, which slows convergence and requires extensive computation. Additionally, DNNs often present complex posterior landscapes that these methods struggle to navigate, leading to memory constraints and convergence issues. To overcome these limitations, ensemble-based techniques have emerged as a promising alternative, offering scalable and efficient approaches to capture epistemic uncertainty. In this study, we employ the Stochastic Weight Averaging Gaussian (SWAG)~\cite{maddox2019simple}, in combination with a DeepEnsemble framework~\cite{lakshminarayanan2017simple}, to approximate the posterior $p(\boldsymbol{\theta}|\mathcal{D})$.  This approach is grounded in the hypothesis by Mandt et al.~\cite{mandt2017stochastic}, suggesting that stochastic gradient descent (SGD) iterations can empirically approximate a Markov chain of samples with a stationary distribution. Here, an SGD trajectory is interpreted as an approximate Bayesian MC sampler, effectively estimating the local geometry of the posterior distribution of the trainable parameters $\boldsymbol{\theta}$. To enhance scalability and simplify the process, the local posterior geometry is characterized primarily by its first and second moments. This involves collecting only the running time mean and variance of $\boldsymbol{\theta}$ along the SGD trajectory to form a Gaussian approximation of the local posterior, a method known as SWAG. Given the complex and multimodal nature of DNN posterior landscapes, SWAG is ideally integrated within the DeepEnsemble framework. As illustrated in Fig.~\ref{fig:SWAG_pred}, this approach entails independently obtaining an ensemble of SGD trajectories in parallel, with each trajectory potentially approximating different peaks of the posterior. This method not only simplifies the representation of the posterior but also ensures a comprehensive exploration of the model’s parameter space, significantly enhancing the accuracy and reliability of epistemic uncertainty quantification in hybrid neural models.

Specifically, the training involves multiple ($N_m$) instances of the DiffHybrid model, each initialized with a distinct set of initial parameters $\boldsymbol{\theta}^{(j)}_0, j = 1, \cdots, N_m$. These models are independently updated using the SWAG algorithm. For each SGD trajectory $\boldsymbol{\theta}^{(j)}_{traj} = \big\{\boldsymbol{\theta}^{(j)}_i\big\}_{i=1}^{T}$, the Gaussian statistics $\big\{ \boldsymbol{\theta}^{(j)}_{\text{SWA}}, \Sigma^{(j)}_{\text{diag}}, \hat{D}^{(j)}\big\}$ are computed on the fly. Here $\boldsymbol{\theta}^{(j)}_{\text{SWA}}$ is the mean, $\Sigma^{(j)}_{\text{diag}}$ is diagonal covariance matrix, and $\hat{D}^{(j)} = \boldsymbol{\theta}^{(j)}_{traj} - \boldsymbol{\theta}^{(j)}_{\text{SWA}}$ is the deviation matrix for given $\boldsymbol{\theta}^{(j)}_{traj}$, respectively. 
To limit the rank and reduce the memory usage, $\hat{D}^{(j)}$ is often estimated by retaining only the last $r$ vectors of $\hat{D}^{(j)}$ corresponding to the last $r$ samples obtained during training, thereby the deviation matrix results in $\hat{D}^{(j)}=[\textbf{d}_{T-r+1}^{(j)},...,\textbf{d}_{T}^{(j)}]$ with column vector $\textbf{d}_k^{(j)} = \boldsymbol{\theta}_k^{(j)} - \boldsymbol{\theta}^{(j)}_{\text{SWA}}$. $T$ represents the total number of samples in each SGD trajectory. Notably, the sampling can be conducted intermittently, every few epochs, with the interval being adjustable. Consequently, the trajectory length $T$ can be less than the total number of training epochs. Additionally, to maintain memory efficiency, only the key statistics, rather than the entire SGD trajectory $\boldsymbol{\theta}_{traj}^{(j)}$, are stored during training. These treatments ensure the scalability and computational feasibility for large DL and hybrid neural models. Upon the completion of SWAG training, the local geometry of the posterior $p(\boldsymbol{\theta}|\mathcal{D})$ is approximated by the Gaussian density defined by each set of $\big\{ \boldsymbol{\theta}^{(j)}_{\text{SWA}}, \Sigma^{(j)}_{\text{diag}}, \hat{D}^{(j)}\big\}$. The local posterior can be sampled using the following equation,
\begin{equation}
    \tilde{\boldsymbol{\theta}}^{(j)} = \boldsymbol{\theta}_{\text{SWA}}^{(j)} + \frac{1}{\sqrt{2}} \cdot \Sigma_{\text{diag}}^{(j)} z_1 + \frac{1}{\sqrt{2(r-1)}}\hat{D}^{(j)}z_2, \quad z_1 \sim \mathcal{N}(0,I_d), z_2 \sim \mathcal{N}(0, I_r),
\end{equation}
where $d$ represents the number of training parameters in $\boldsymbol{\theta}^{(j)}$, and $I_d$ and $I_r$ are two identity matrices of ranks $d$ and $r$, respectively. Using the idea of DeepEnsemble method, the full posterior distribution can be approximated by aggregating the Gaussian samples from multiple independent SWAG local posterior approximations, $\big\{ \boldsymbol{\theta}^{(j)}_{\text{SWA}}, \Sigma^{(j)}_{\text{diag}}, \hat{D}^{(j)} \big\}_{j=1}^{N_m}$. SWAG effectively captures the posterior geometry around local minima, while the ensemble approach addresses the multimodal nature of the posterior, thereby facilitating the efficient estimation of epistemic/model uncertainty. Consequently, our approach involves training multiple DiffHybrid model instances with SWAG simultaneously to capture different Gaussian posterior distributions characterized by low-rank covariance matrices. The overall training algorithm of the proposed DiffHybrid-UQ model is detailed in Algorithm~\ref{alg:DiffHybrid-UQ-train} and the SWAG implementation details are provided in \ref{sec:SWAG}. 
\begin{algorithm}[hbt!]
{
\caption{Training algorithm for DiffHybrid-UQ model}\label{alg:DiffHybrid-UQ-train}
\KwData{Read experimental data $\mathcal{D}=\{x_i,\mathbf{v}_i\}^N_{j=1}$}
\textbf{Initialize:} Initialize trainable parameters $\{\boldsymbol{\theta}_0^{(j)}\}_{j=1}^{N_m} = \{\boldsymbol{\theta}_0^{(1)}, \boldsymbol{\theta}_0^{(2)}, \ldots , \boldsymbol{\theta}_0^{(N_m)}\}$ \\
\For{i = 1 to N\_epoch}
{   \Comment {Training multiple models with different $\boldsymbol{\theta}^{(j)}$}\\
    \For{$\boldsymbol{\theta}_i^{(j)} = \{\boldsymbol{\theta}_i^{(j)}\}_{j=1}^{N_m}$} 
    {
    $(\mathbf{v}_{\theta}, \Sigma^2_{V,\theta}) \gets \text{DIffHybrid-UQ} \big(\mathbf{V}_{init}=(\mathbf{v}_0, \boldsymbol{\theta}^{(j)}_{i,\Sigma^2_0})),  \boldsymbol{\theta}_i^{(j)} = (\boldsymbol{\theta}^{(j)}_{i,\lambda_{U}}, \boldsymbol{\theta}^{(j)}_{i,\mathcal{NN}}, \boldsymbol{\theta}^{(j)}_{i,\Sigma^2_0}) \big)$\\
    $(\mathbf{v}^{(p)}_{\theta}, {\Sigma^2_{V,\theta}}^{(p)})  \gets \text{Map\_to\_Observable}(\mathbf{v}_{\theta}, \Sigma^2_{V,\theta})$ \Comment {Exp observables}\\
    $\mathcal{L} \gets \frac{log({\Sigma^2_{V,\theta}}^{(p)})}{2} + \frac{(V_i - \mathbf{v}^{(p)}_{\theta})^2}{2{\Sigma^2_{V,\theta}}^{(p)}}$ \Comment {Loss function}\\
    $\boldsymbol{\theta}_{i+1}^{(j)} \gets \boldsymbol{\theta}_i^{(j)} - \lambda\nabla_{\theta} \mathcal{L}(\boldsymbol{\theta}_i^{(j)}) $\Comment {Update $\boldsymbol{\theta}^{(j)}$ parameters}\\
        \If {MOD(i, interval) = 0}
        {
        $n \gets i/interval$\\
        $\bar{\boldsymbol{\theta}}^{(j)} \gets \frac{n\bar{\boldsymbol{\theta}}^{(j)}+\boldsymbol{\theta}_{i+1}^{(j)}}{n+1}$\Comment {Update movements}\\
        $\bar{\boldsymbol{\theta}^2}^{(j)} \gets \frac{n\bar{\boldsymbol{\theta}^2}^{(j)}+{\boldsymbol{\theta}^2_{i+1}}^{(j)}}{n+1}$\\
        $\hat{D}^{(j)} \gets [\textbf{d}_{i-r}^{(j)},...,\textbf{d}_{i+1}^{(j)}] $ \Comment {Store deviation}
        } 
    }
}
}
\end{algorithm}

\section{Results}
\label{sec:result}
This section presents extensive numerical experiments conducted to assess the performance of the proposed DiffHybrid-UQ framework in modeling various dynamical systems governed by ODEs and PDEs, with a focus on uncertainty quantification.

\subsection{Modeling dynamical systems governed by ODEs}

We first study the proposed DiffHybrid-UQ model on the Hamiltonian systems governed by the simple ODEs described as follows,
\begin{subequations}
\begin{alignat}{2}
    \frac{\partial x_1}{\partial t} &= f_1(x_1,x_2) = x_2,\\
    \frac{\partial x_2}{\partial t} &= f_2(x_1,x_2) = -sin(x_1), 
\end{alignat}
\label{eq:ham_ode}
\end{subequations}
where $\mathbf{v} = [x_1, x_2]^T \in \mathbb{R}^2$ is a two-dimensional state variable, and the functions $f_1, f_2: \mathbb{R}^2 \Rightarrow \mathbb{R}^1 $ can be linear or nonlinear, defining the right-hand side of the governing equations. Synthetic data are generated by solving Eq.~\ref{eq:ham_ode}, and Gaussian random noise is added to simulate measurement uncertainties typical of real-world experimental data. To illustrate the DiffHybrid-UQ model's capabilities, the governing physics is assumed to be only partially known. Specifically, while the function $f_2$ is known, $f_1$ is considered unknown. Therefore, we construct an auto-regressive hybrid neural model based on Eq.~\ref{eq:ham_ode}, where a trainable Bayesian neural network (BNN), $\mathcal{U}_{nn}: \mathbb{R}^4 \Rightarrow \mathbb{R}^2$ represents the right-hand side of Eq.~\ref{eq:ham_ode}a, processing the mean and variance of $\mathbf{v}$ to predict the Gaussian statistics of $f_1$. The known function $f_2$ is encapsulated within a UT function, resulting a probabilistic function, $\mathcal{K} = UT(f_2): \mathbb{R}^4 \Rightarrow \mathbb{R}^2$. Both $\mathcal{U}_{nn}$ and $\mathcal{K}$ function as probabilistic entities, handling random variables to capture the inherent uncertainties in the training data. 

During the training phase, the DiffHybrid-UQ model is trained with data spanning a single trajectory over the time interval $[0:\Delta t:T^{'}]$, where $T^{'}<T$. The model's predictive performance is assessed in the forecast/testing range $(T^{'}:\Delta t:T]$. To reflect realistic experimental conditions, different measurement errors are introduced for different variables, acknowledging that measurement uncertainty can differ across variables. To intensively assess the model's capacity for uncertainty prediction, we conduct three test cases:
\begin{table}[h]
    \centering
    \begin{tabular}{c c c}
        \hline
        Case & Data $\mathcal{D}$ & Description \\
        \hline
        1 & $(x_1, x_2 \in \mathbb{R}^1 \times [0:0.1:20 s])$ & Complete data for both variables are provided\\
        2 & $(x_1, x_2 \in \mathbb{R}^1 \times [5:0.1:20 s])$ & Partial data for both variables are provided\\
        3 & $(x_2 \in \mathbb{R}^1 \times [0:0.1:20 s])$ & Partial data for only one variable is provided\\
        \hline
    \end{tabular}
    \label{tab:my_label}
\end{table}\vspace{-1em}
These scenarios serve to test the model's efficacy in accurately estimating uncertainties under varying data availability conditions. In-depth analyses and results of these test cases will be discussed in subsequent sections.

\subsubsection{Case 1: Complete data for both state variables}
\begin{figure}[!ht]
    \centering
    \includegraphics[width=\linewidth]{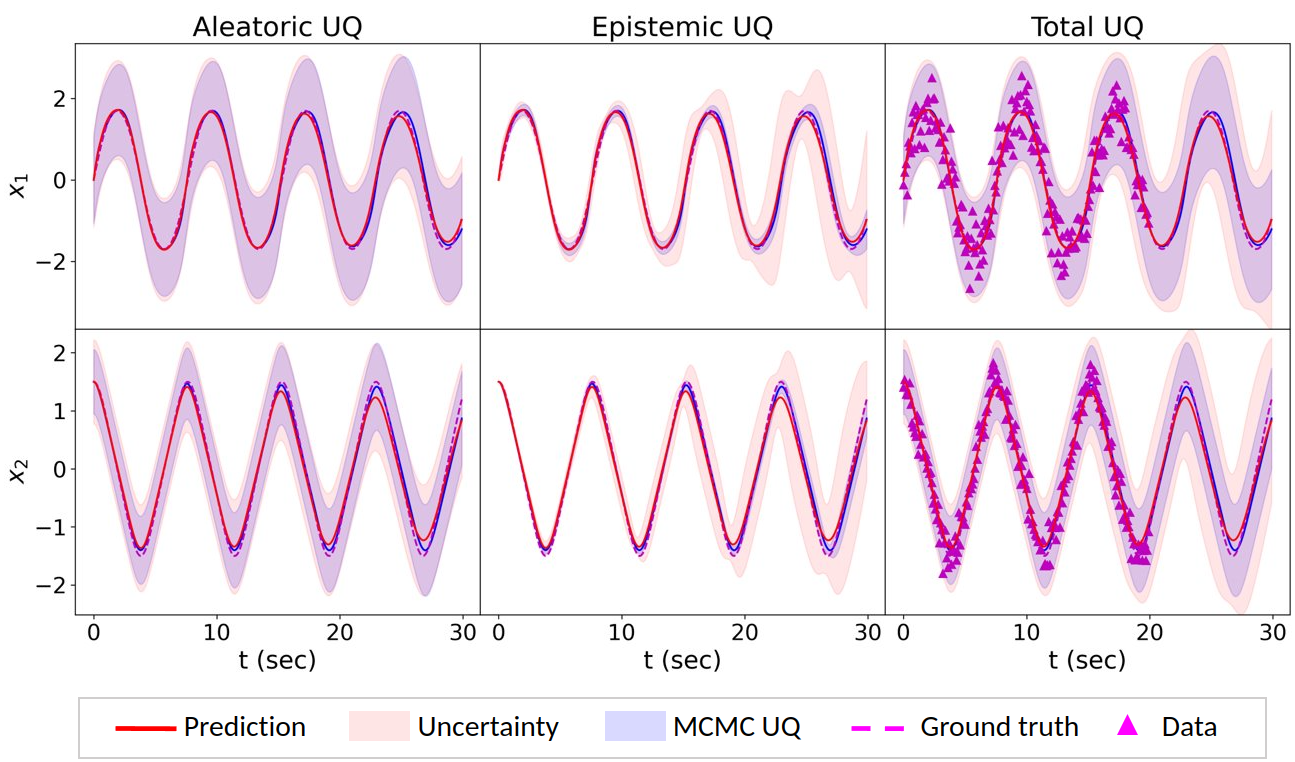}
    \caption{Compares the DiffHybrid-UQ model's prediction and UQ against HMC method. Training data spans $\mathcal{D} = (x_1, x_2 \in \mathbb{R}^1 \times [0:0.1:20 s])$ with testing in the $(x_1, x_2 \in \mathbb{R}^1 \times (20:0.1:30 s])$ region.}
    \label{fig:pred_C0_Ham}
\end{figure}

The DiffHybrid-UQ model is trained using data for both the $x_1$ and $x_2$ variables over 20-second interval $\mathcal{D} = (x_1, x_2 \in \mathbb{R}^1 \times [0:0.1:20 s])$. The training, spanning 10,000 epochs, yields results depicted in Fig.~\ref{fig:pred_C0_Ham}. To verify the UQ performance of the proposed DiffHybrid-UQ model, we also perform exact Bayesian learning using Hamiltonian Monte Carlo (HMC) method, often considered the gold standard for Bayesian UQ. For HMC, 150,000 samples are obtained, initialized at the local minima obtained from gradient descent training, thereby bypassing the burn-in phase to enhance computational efficiency. A side-by-side comparison of the DiffHybrid-UQ and HMC predictions is shown in Fig.~\ref{fig:pred_C0_Ham}, with color-shaded regions representing 3-STD confidence intervals. 
The prediction mean of the DiffHybrid model aligns very well with the ground truth for both $x_1$ and $x_2$, even in the forecasting region $(20:0.1:30 s]$, demonstrating model's forecasting capability. Overall, the total prediction uncertainty of the DiffHbrid-UQ model matches that of the HMC method. Notably, the shaded areas in the first column, indicative of aleatoric uncertainty, differ for variables $x_1$ and $x_2$, corresponding to the varying noise levels in the data, as shown in the third column. The aleatoric uncertainty estimated by the DiffHybrid-UQ model closely agree with that of the HMC method, reflecting its effectiveness in capturing inherent data uncertainty. Regarding epistemic uncertainty, represented by shaded areas in the second column, these remain small but increase in the forecast region where data are unavailable. This rise in epistemic uncertainty is expected and reasonable due to the lack of data in this region. However, a noticeable difference in epistemic uncertainty predictions between the DiffHybrid-UQ and HMC methods exists, likely due to the initialization of all HMC samples at a single local minimum, in contrast to the DeepEnsemble strategy of the DiffHybrid-UQ model, which more effectively explores the multimodal posterior.

\begin{figure}[!ht]
    \centering
    \subfloat{\includegraphics[width=\linewidth]{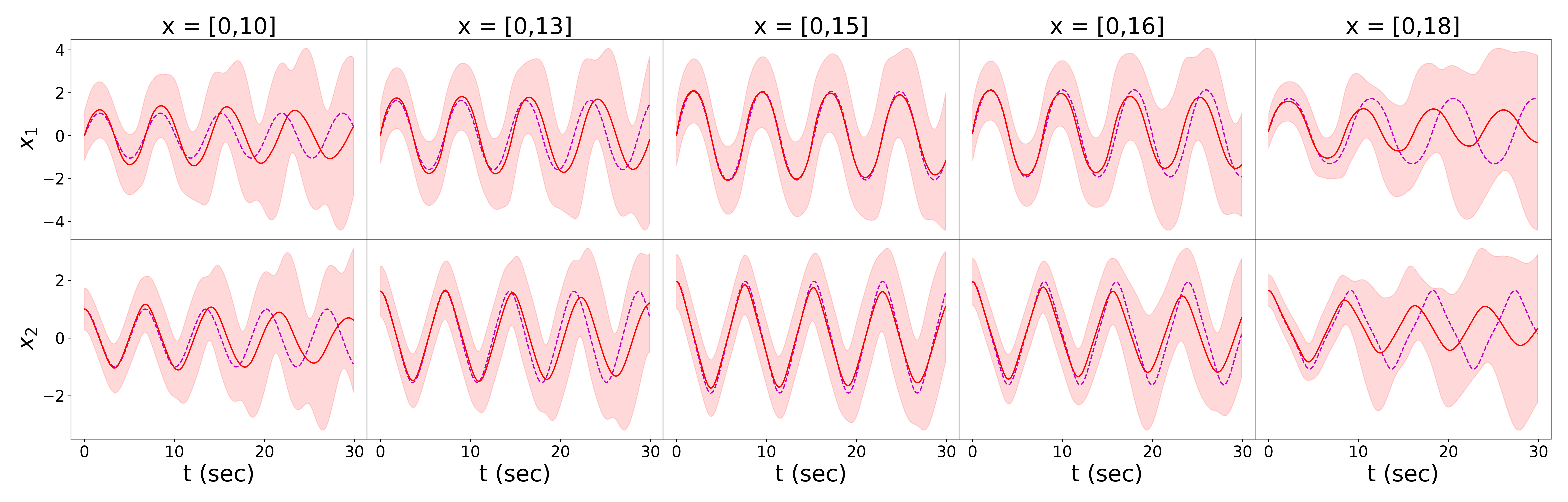}}\\
    \subfloat{\includegraphics[width=0.5\linewidth]{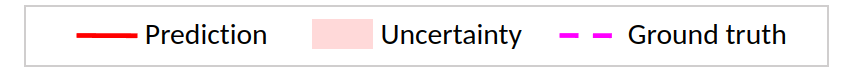}}
    \caption{DiffHybrid-UQ prediction with quantified uncertainties compared against the ground truth for different testing initial conditions. Trained initial condition: $\mathbf{x}=[0,15]$.}
    \label{fig:pred_C0_Ham_test}
\end{figure}

To evaluate the generalization capability of the DiffHybrid learning framework, the trained model is tested on various trajectories generated under different initial conditions. Figure~\ref{fig:pred_C0_Ham_test} shows how the model predictions compared with the ground truth across these scenarios.
As predictions deviate from the training trajectory, initialized at $\mathbf{x}=[0,15]$, there is a notable decrease in mean prediction accuracy, accompanied by increased uncertainty. This trend, evident in the parametric extrapolation region, indicates the DiffHybrid-UQ model is able to reasonably estimate the confidence of its prediction, which diminish as it moves beyond the range of the training scenarios. 

Overall, the DiffHybrid-UQ model exhibit promising performance, producing uncertainty estimates that are in line with those obtained from the HMC samping.  Its ability to quantify both aleatoric and epistemic uncertainties offers valuable insights into the reliability of its predictions, effectively addressing the inherent uncertainties stemming from both data variability and model formulation.

\subsubsection{Case 2: Sparse data for both variables}
\begin{figure}[!ht]
    \centering
    \subfloat{\includegraphics[width=0.9\textwidth]{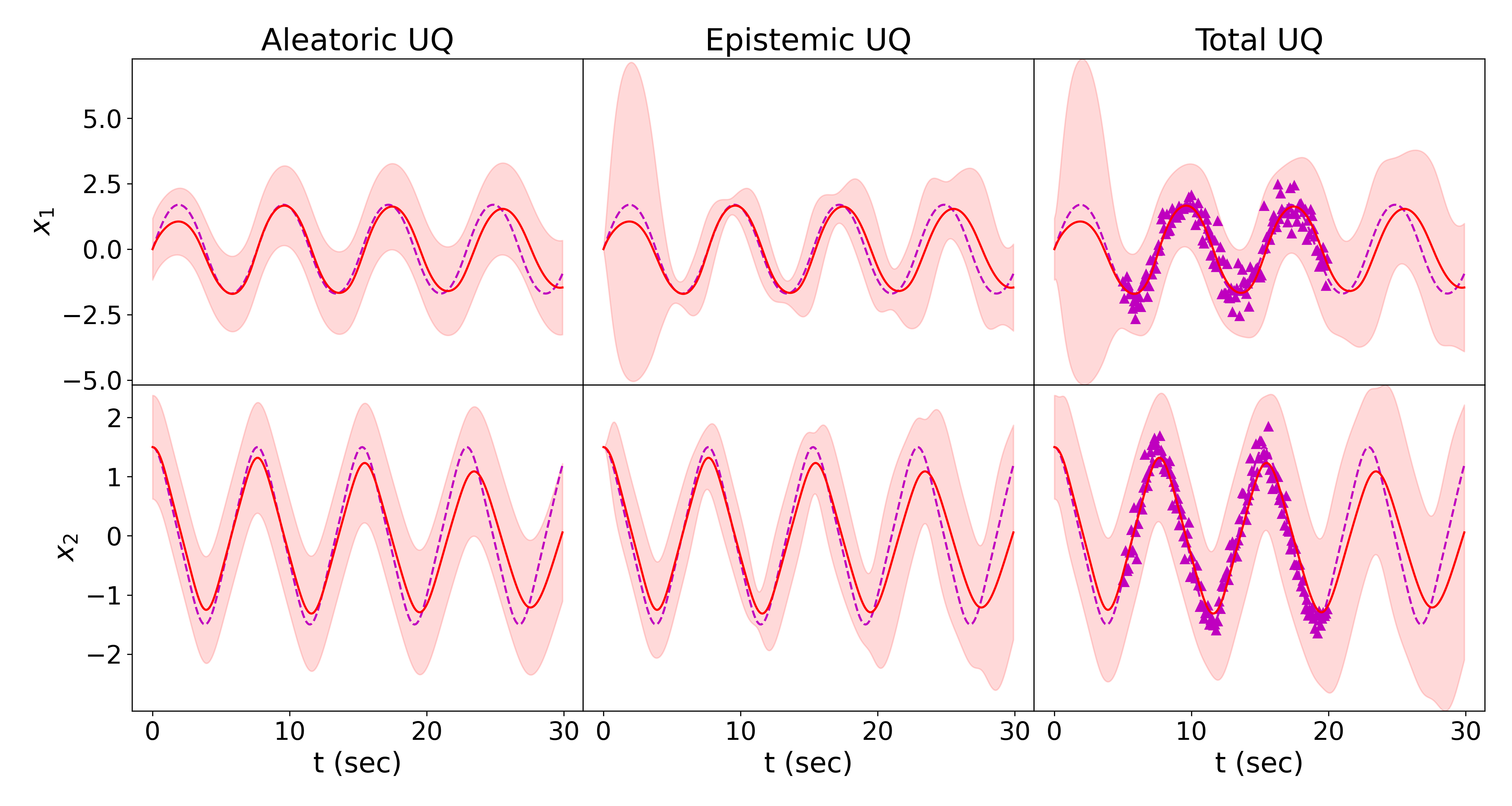}}\\
    \subfloat{\includegraphics[width=0.6\textwidth]{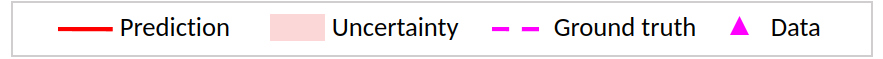}}
    \caption{Comparison of the DiffHybrid-UQ model’s prediction with UQ against the ground truth.
    Training data is $\mathcal{D} = (x_1, x_2 \in \mathbb{R}^1 \times [5:0.1:20 s])$, with testing in $(x_1, x_2 \in \mathbb{R}^1 \times [0:0.1:5 s) \cup (20:0.1:30 s])$. }
    \label{fig:pred_C1_Ham}
\end{figure}

In this scenario, the DiffHybrid-UQ model is trained with partial data for both the $x_1$ and $x_2$ variables, specifically within the time range of $[5:0.1:20 s]$, denoted as $\mathcal{D} = (x_1, x_2 \in \mathbb{R}^1 \times [5:0.1:20 s])$. Notably, data for the initial 5 seconds is not included. The model undergoes a training period of 10,000 epochs, with results presented in Fig.~\ref{fig:pred_C1_Ham}. 
The confidence intervals of the model predictions effectively reflect the data uncertainty for both $x_1$ and $x_2$. While the prediction for $x_1$ shows a notable deviation from the ground truth, the model aptly indicates lower confidence in this region. On the other hand, for $x_2$, the model's uncertainty remains relatively low in the same interval, suggesting a good alignment between the DiffHybrid-UQ predictions and the ground truth. The model's performance in unseen data regions is reasonable, with an increasing level of uncertainty observed towards the forecasting region $(20:0.1:30 s]$. Thus, the DiffHybrid-UQ model showcases its ability to estimate uncertainties effectively, even with partially available data. By precisely capturing data uncertainties and indicating reduced confidence in data-sparse regions, the model offers valuable insights into the reliability of its predictions.

\subsubsection{Case 3: Incomplete data for only one state variable}

\begin{figure}[!ht]
    \centering
    \subfloat{\includegraphics[width=\linewidth]{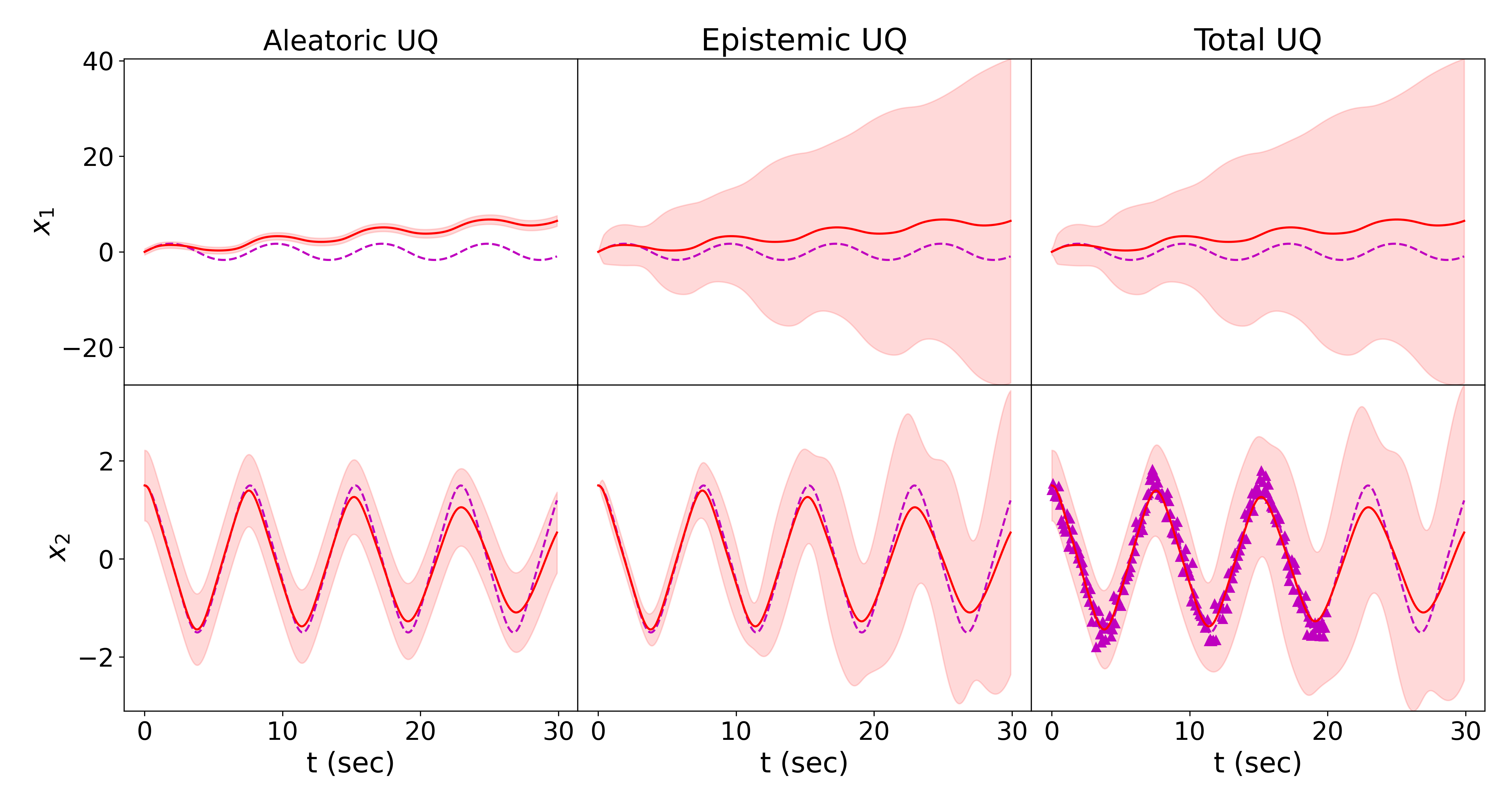}}\\
    \subfloat{\includegraphics[width=0.6\textwidth]{./fig/label_pugd.png}}
    \caption{Comparison of the DiffHybrid-UQ model’s prediction with UQ against the ground truth.
    Training data is $\mathcal{D} = ({x}_2 \in \mathbb{R}^1 \times [0:0.1:20s])$, with 
    testing in $({x}_1 \in \mathbb{R}^1 \times [0:0.1:30 s], {x}_2 \in \mathbb{R}^1 \times (20:0.1:30 s])$. }
    \label{fig:pred_C2_Ham}
\end{figure}

A key strength of hybrid neural models lies in their ability to be trained with incomplete data, even when labels for certain state variables are absent. To demonstrate this capability, we trained the DiffHybrid-UQ model using only the data for $x_2$, designated as  $\mathcal{D} = ({x}_2 \in \mathbb{R}^1 \times [0:0.1:20s])$. This training, encompassing 10,000 epochs, produced results illustrated in Fig.~\ref{fig:pred_C2_Ham}.
The results show that the DiffHybrid-UQ model effectively captures the dynamics of $x_2$, while the $x_1$ prediction mean notably deviates from the ground truth. This is due to that the information for $x_1$ is learned indirectly from the labeled data of $x_2$. For $x_1$, the model understandably exhibits much higher uncertainty as not directly labels are provided. In contrast, for $x_2$, the model displays relatively low uncertainty. In line with previous findings, confidence interval increases in the forecast region, reflecting the increased model uncertainty beyond the training data.  Its ability to reflect data scarcity and increase uncertainty in unobserved regions highlights its robustness in limited data scenarios.

\subsection{Modeling spatiotemporal dynamics governed by PDEs}

In this section, the DiffHybrid-UQ is utilized to model spatiotemporal dynamics with quantified uncertainty, which are also spatiotemporal fields. Specifically, its efficacy is assessed on a reaction-diffusion system governed by the following PDEs,  
\begin{subequations}
\begin{alignat}{2}
    \dot{v}_1 = D_1 \nabla^2 v_1 + s_1(v_1,v_2) \\ 
    \dot{v}_2 = D_2 \nabla^2 v_2 + s_2(v_1,v_2)
\end{alignat}
\end{subequations}
where $v_1, v_2  \in \mathbb{R}^{n_x \times n_y}$ are the state variables over a discretized 2-D domain, characterized by $n_x \times n_y$ grid points in the $x, y$ directions. The diffusion coefficients are set as $D_1=2.8\times10^{-4}$ and $D_2=5.0\times10^{-2}$. The source terms $s_1(v_1,v_2)=v_1-v_1^3-v_2-0.005$ and $s_2(v_1,v_2)=10(v_1-v_2)$ define the reaction dynamics. Neumann boundary conditions are applied in these simulations. Similar to ODE cases above, synthetic datasets for these PDEs are generated and perturbed with Gaussian random noises. We assume that the model forms for the reaction dynamics is partially known: $s_1$ is given, while $s_2(v_1,v_2)$ remains unknown. Therefore, in our DiffHybrid-UQ framework, the known $s_1(v_1,v_2)$ is encapsulated within an UT function, resulting in a known random function $\mathcal{K}^(V) = UT(s_1)(V): \mathbb{R}^4 \Rightarrow \mathbb{R}^2$, which is seamlessly integrated with Bayesian neural networks $\mathcal{U}_{bnn}(V): \mathbb{R}^4 \Rightarrow \mathbb{R}^2$ to learn the missing reaction physics. The diffusion operator is encoded as known convolution operations using finite difference method (FDM),
\begin{equation}
    \nabla^2 v(i,j) = \frac{v(i+1,j) + v(i-1,j) - 2v(i,j)}{\Delta x^2} + \frac{v(i,j+1) + v(i,j-1) - 2v}{\Delta y^2}.
\end{equation}
where $v(i+1,j), v(i-1,j), v(i,j+1), v(I,j-1)$ are neighbouring values of $v(i,j)$ post-discretization. Now, the discretized diffusion term as a linear combination of adjacent spatial points leads to the output mean and variance as,
\begin{subequations}
\begin{alignat}{2}
    \mu_{\nabla^2 V}(i,j) = \frac{V(i+1,j) + V(i-1,j) - 2V(i,j)}{\Delta x^2} + \frac{V(i,j+1) + V(i,j-1) - 2V}{\Delta y^2} \\ 
    \Sigma^2_{\nabla^2 V}(i,j) = \frac{\Sigma^2_{V(i+1,j)} + \Sigma^2_{V(i-1,j)} + 4\Sigma^2_{V(i,j)}}{\Delta x^4} + \frac{\Sigma^2_{V(i,j+1)} + \Sigma^2_{V(i,j-1)} + 4\Sigma^2_{V(i,j)}}{\Delta y^4}
\end{alignat}
\label{eq:LapFDM}
\end{subequations}
Furthermore, diffusion coefficients $D_i$ are considered as unknown variables and kept trainable throughout all test cases, unless specifically mentioned otherwise. The Hybrid models are typically employed in scenarios with limited, indirect data. To replicate such conditions, unless otherwise stated, we provide training data only for the $v_2$ variable and exclude any labels for $v_1$. Training is conducted on a specific trajectory within the timeframe $[0:\Delta t:T^{'}]$, thus $\mathcal{D} = ({v}_2 \in \mathbb{R}^{n_x \times n_y} \times [0:\Delta t:T^{'}])$, where $T^{'} < T$. The model's predictive performance is then evaluated in the test region $({v}_1 \in \mathbb{R}^{n_x \times n_y} \times [0:\Delta t:T], v_2 \in \mathbb{R}^{n_x \times n_y} \times (T^{'}:\Delta t:T])$.

\subsubsection{Dynamical forecasting with spatiotemporal uncertainty estimation}

\begin{figure}[!ht]
    \centering
    \subfloat[$v_1$]{\includegraphics[width=0.9\textwidth]{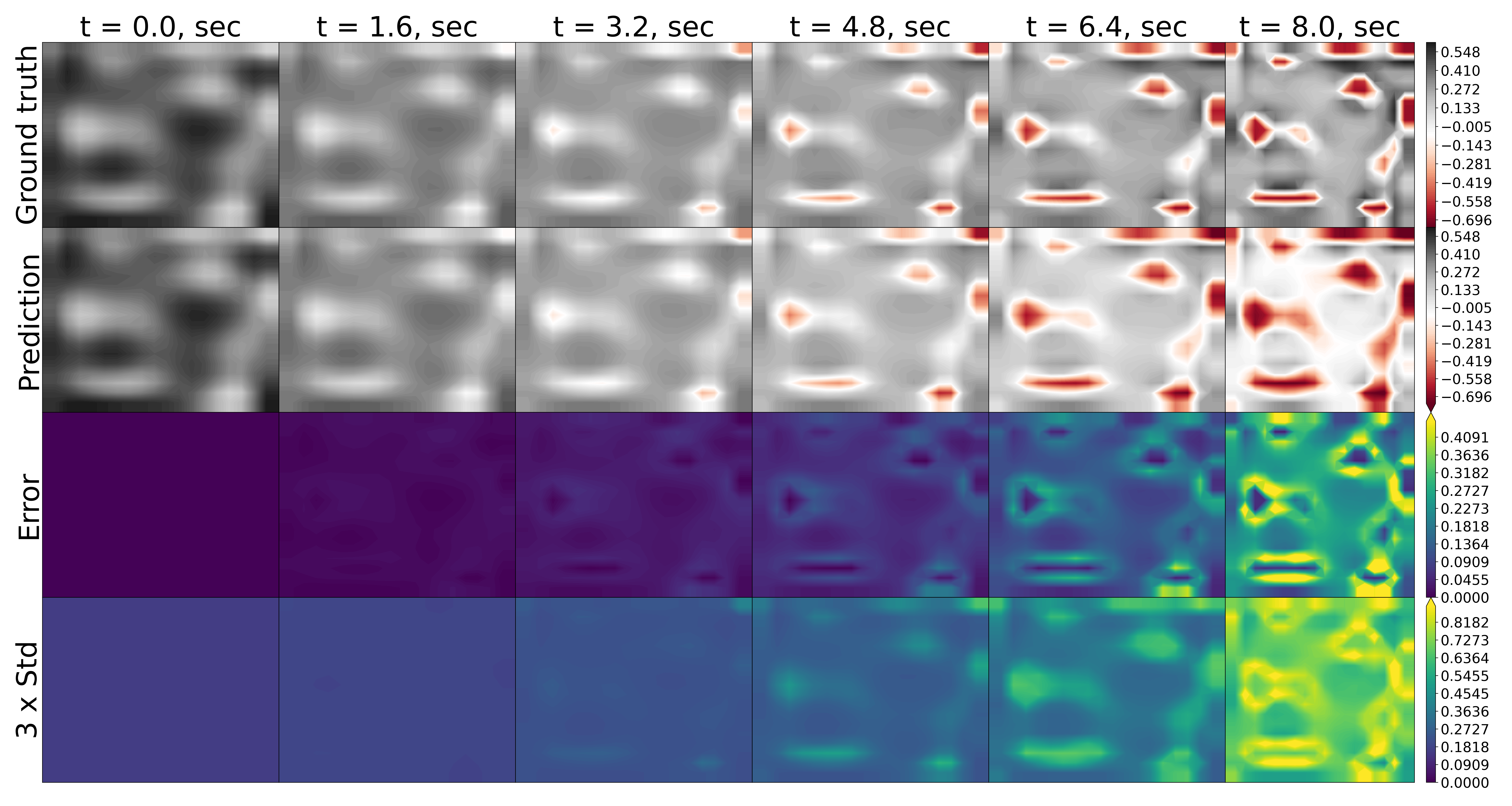}}\\
    \subfloat[$v_2$]{\includegraphics[width=0.9\textwidth]{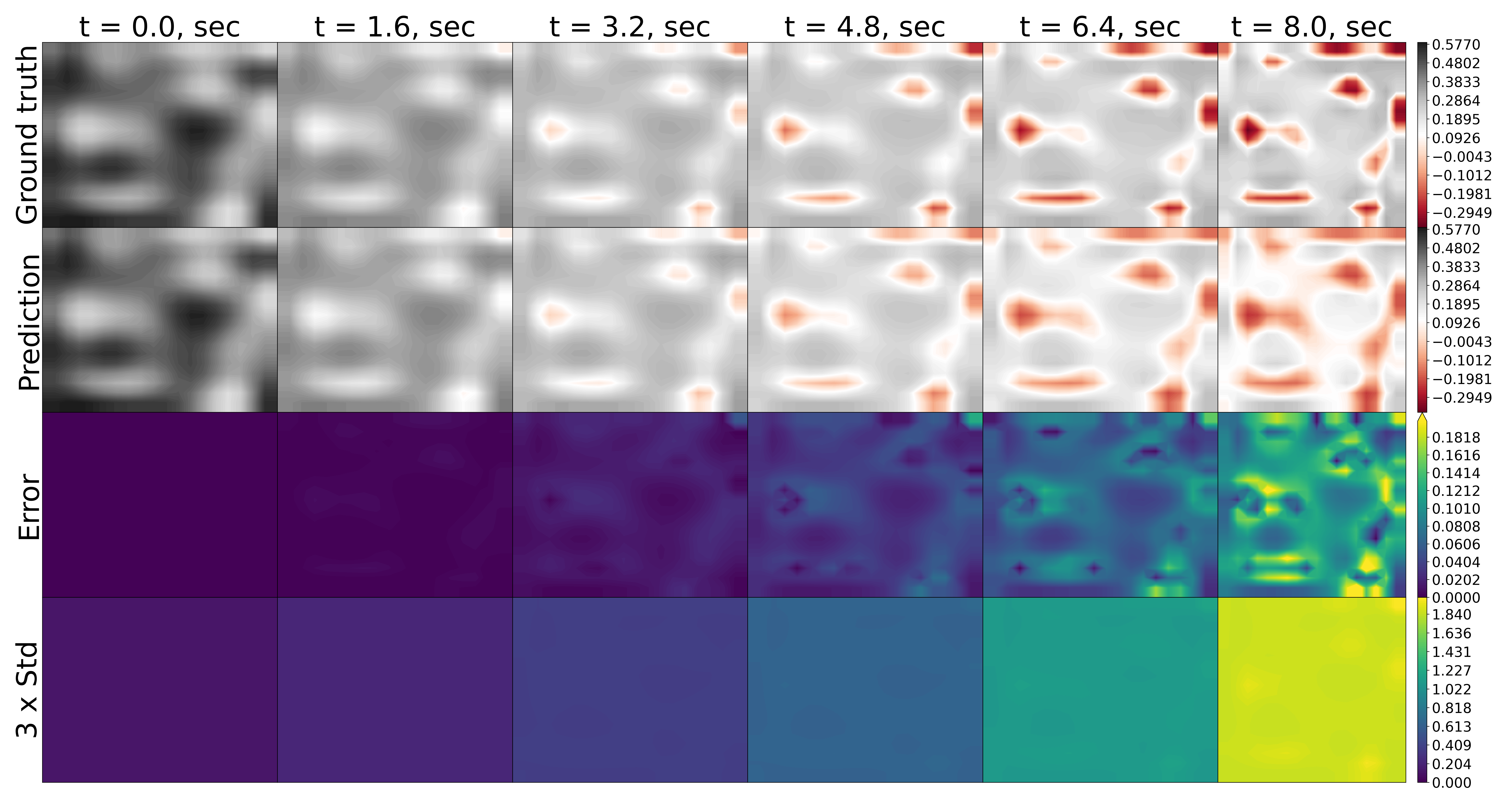}}
    \caption{Comparison of the DiffHybrid-UQ model’s prediction (2nd row) with UQ (4th row) against the ground truth (1st row) for variable $v_1$ (a) and $v_2$ (b) over time. Training data: $\mathcal{D} = ({v}_2 \in \mathbb{R}^{20 \times 20} \times [0:0.01:4 s])$, with testing 
    in $({v}_1 \in \mathbb{R}^{20 \times 20} \times [0:0.01:8 s], {v}_2 \in \mathbb{R}^{20 \times 20} \times (4:0.01:8 s])$. }
    \label{fig:pred_2D_N_GP_train}
\end{figure}

\begin{figure}[!ht]
    \centering
    {\includegraphics[width=0.6\textwidth]{./fig/label_pugd.png}}\\
    \subfloat[$v_1$]{\includegraphics[width=0.5\textwidth]{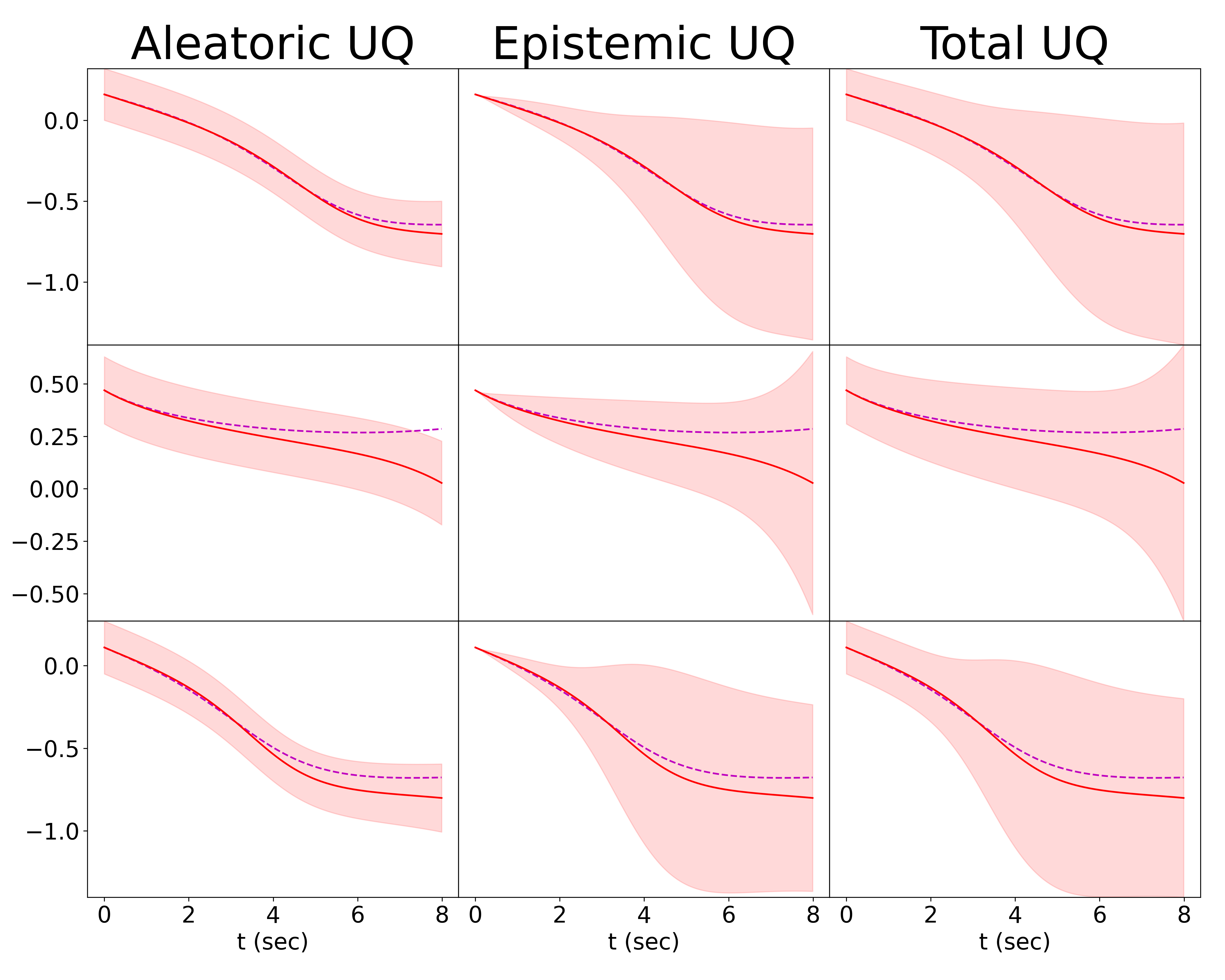}}
    \subfloat[$v_2$]{\includegraphics[width=0.5\textwidth]{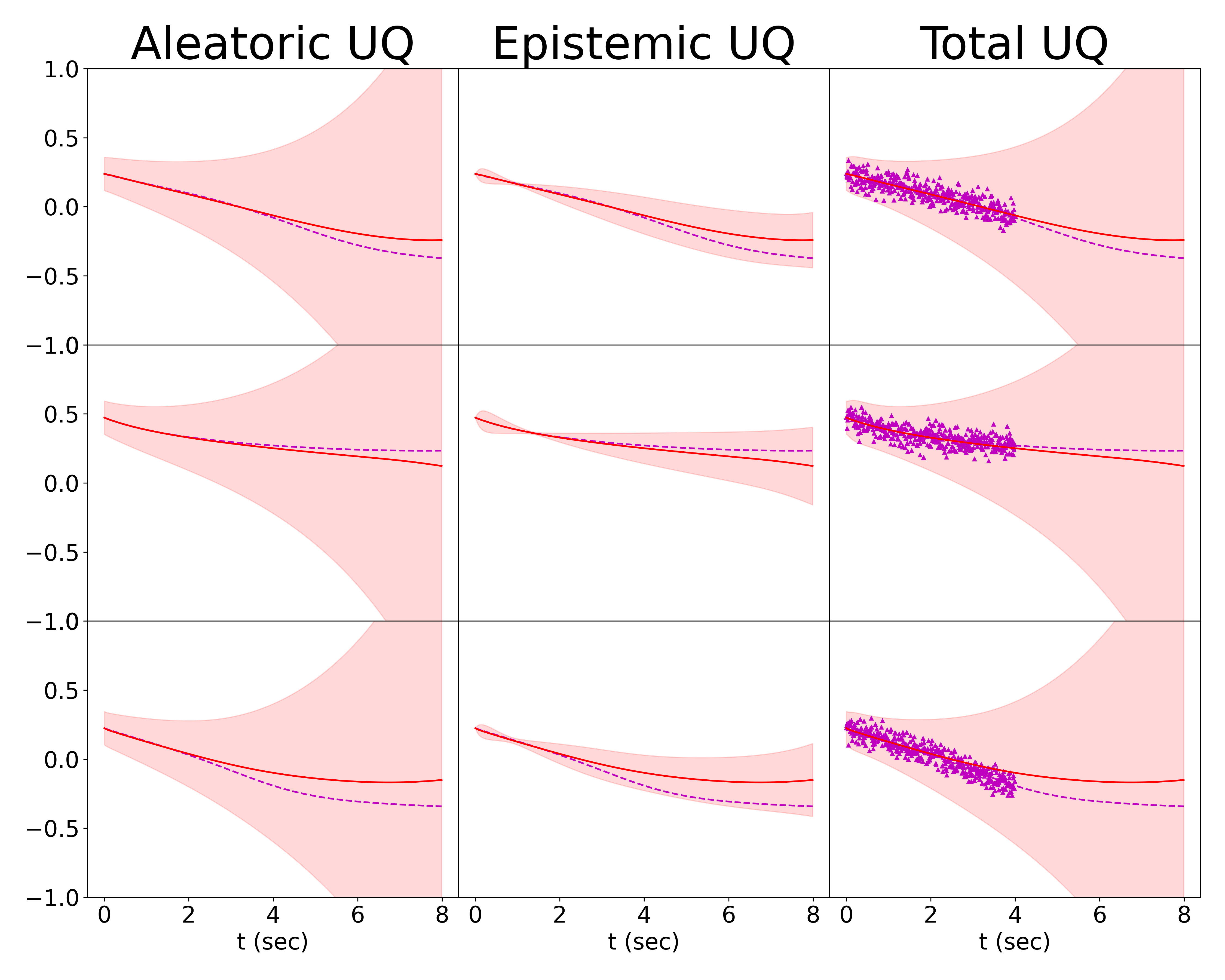}}\\
    \caption{Comparison of the DiffHybrid-UQ model’s prediction with UQ against the ground truth for variable $v_1$ (a) and $v_2$ (b) at three spatial locations over time. Training data: $\mathcal{D} = ({v}_2 \in \mathbb{R}^{20 \times 20} \times [0:0.01:4 s])$, with 
    testing in $({v}_1 \in \mathbb{R}^{20 \times 20} \times [0:0.01:8 s], {v}_2 \in \mathbb{R}^{20 \times 20} \times (4:0.01:8 s])$.}
    \label{fig:pred_1D_N_GP_train}
\end{figure}

To investigate the spatiotemporal behavior of predicted uncertainty, we trained the DiffHybrid-UQ model using the data of a single trajectory for up to 4 seconds on a $20 \times 20$ grid, i.e., $\mathcal{D} = ({v}_2 \in \mathbb{R}^{20 \times 20} \times [0:0.01:4 s])$ and tested it in the forecast region $(4:0.01:8 s]$. For this analysis, the diffusion coefficients $D_i$ were assumed to be known. To clearly observe the spatial uncertainty prediction, the initial condition is defined by a Gaussian random field. The model was trained for 100 epochs, with results displayed in Fig.~\ref{fig:pred_2D_N_GP_train} and Fig.~\ref{fig:pred_1D_N_GP_train}.

As we can see from Fig.~\ref{fig:pred_2D_N_GP_train}, the prediction mean of both $v_1$ and $v_2$ agree with the ground truth reasonably well, while the prediction errors grow temporally. There is a strong correlation between the spatial prediction error (third row) and the estimated uncertainty (fourth row). Notably, no training data was provided for $v_1$, highlighting the model's capability to learn the dynamics and estimate corresponding uncertainty in the absence of direct labels.

Fig.~\ref{fig:pred_1D_N_GP_train} illustrates temporal uncertainty through a shaded region representing three times the standard deviation, signifying the confidence interval. This interval effectively captures the data uncertainty, which increases over time, consistent with the auto-regressive nature of the model where confidence diminishes with each subsequent time-step prediction. The aleatoric uncertainty for $v_1$ remains relatively constant throughout the forecast period since not direct labels are involved for $v_1$. In contrast, the data uncertainty for $v_2$ escalates over time, a consequence of the larger diffusion coefficient $D_2$. Since data for $v_2$ was provided during training, the epistemic uncertainty for $v_2$ is comparatively lower than that for $v_1$. These results affirm that the DiffHybrid-UQ model successfully captures the spatiotemporal behavior of predicted uncertainty.



\subsubsection{Inference and generalization capability of DiffHybrid-UQ model}
\begin{figure}[!t]
    \centering
    \subfloat[$v_1$]{\includegraphics[width=0.9\textwidth]{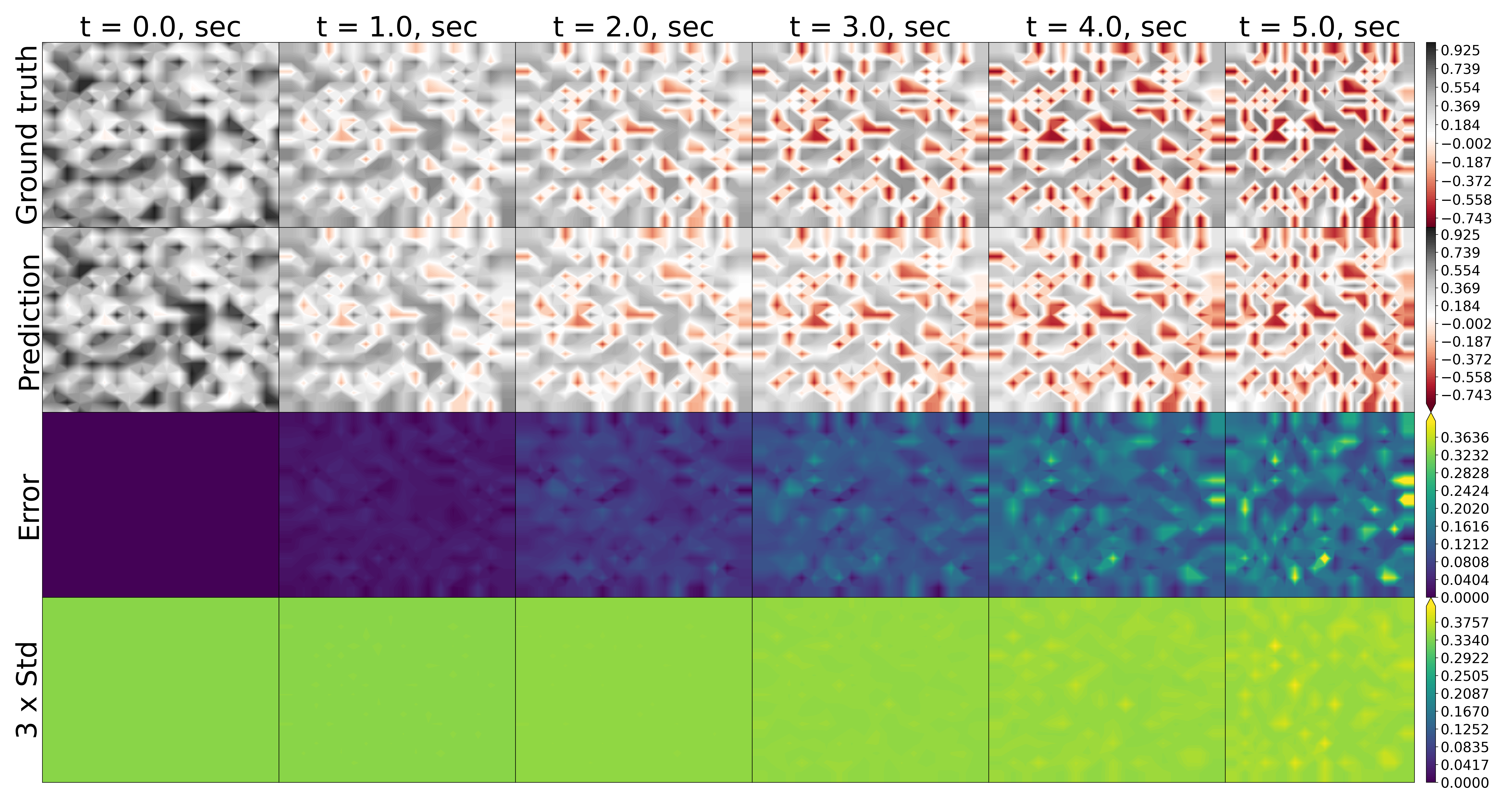}}\\
    \subfloat[$v_2$]{\includegraphics[width=0.9\textwidth]{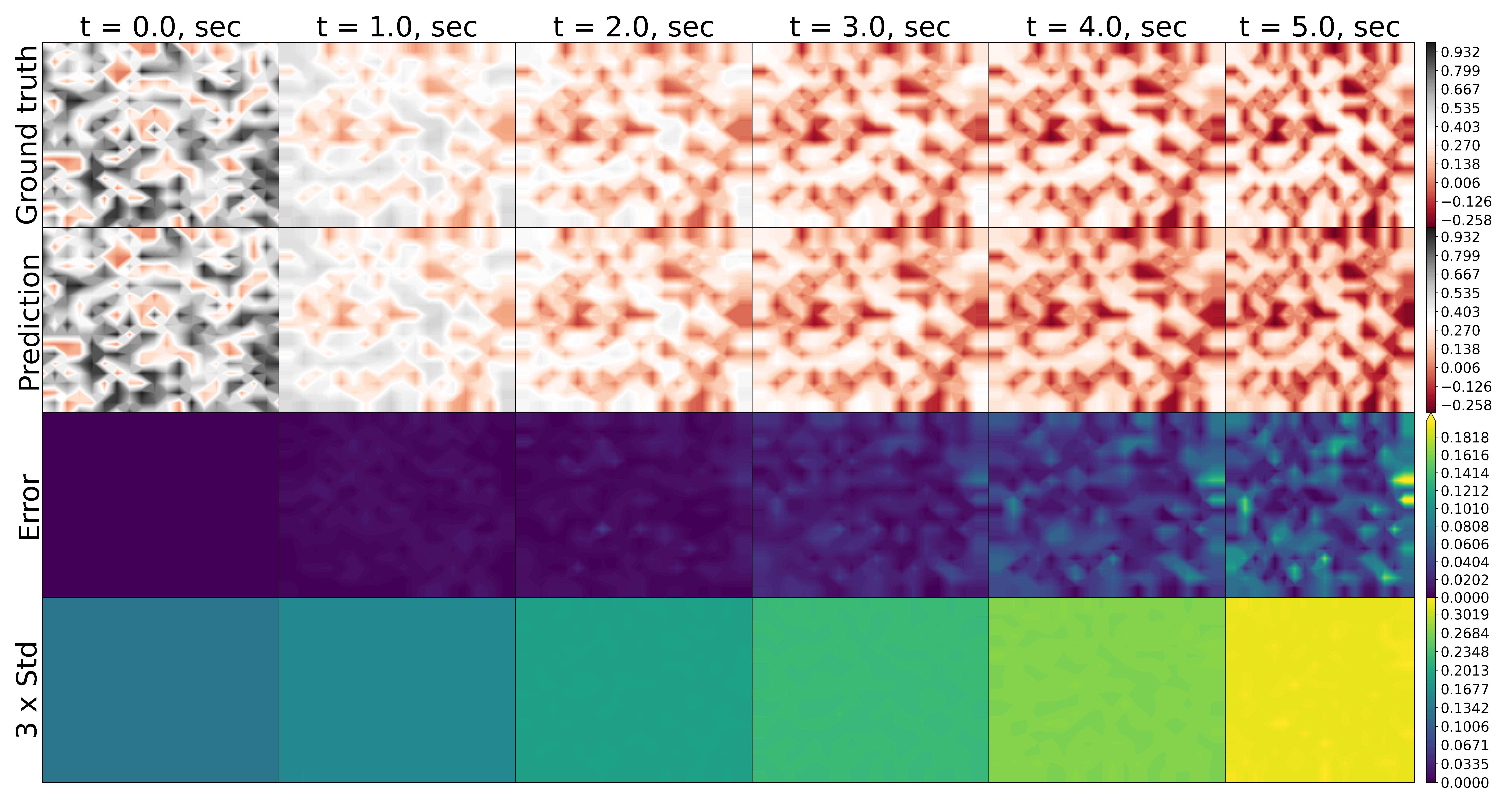}}
    \caption{Comparison of the DiffHybrid-UQ model’s prediction (2nd row) with UQ (4th row) against the ground truth (1st row) for $v_1$ (a) and $v_2$ (b) over time. Training data: $\mathcal{D} = ({v}_2 \in \mathbb{R}^{20 \times 20} \times [0:0.01:3.5 s])$, with 
    testing in $({v}_1 \in \mathbb{R}^{20 \times 20} \times [0:0.01:3.5 s], {v}_2 \in \mathbb{R}^{20 \times 20} \times (3.5:0.01:5 s])$. }
    \label{fig:pred_train_2D_rd_rnd}
\end{figure}

\begin{figure}[!t]
    \centering
     \includegraphics[width=0.6\textwidth]{./fig/label_pugd.png}\\
    \subfloat[$v_1$]{\includegraphics[width=0.5\textwidth]{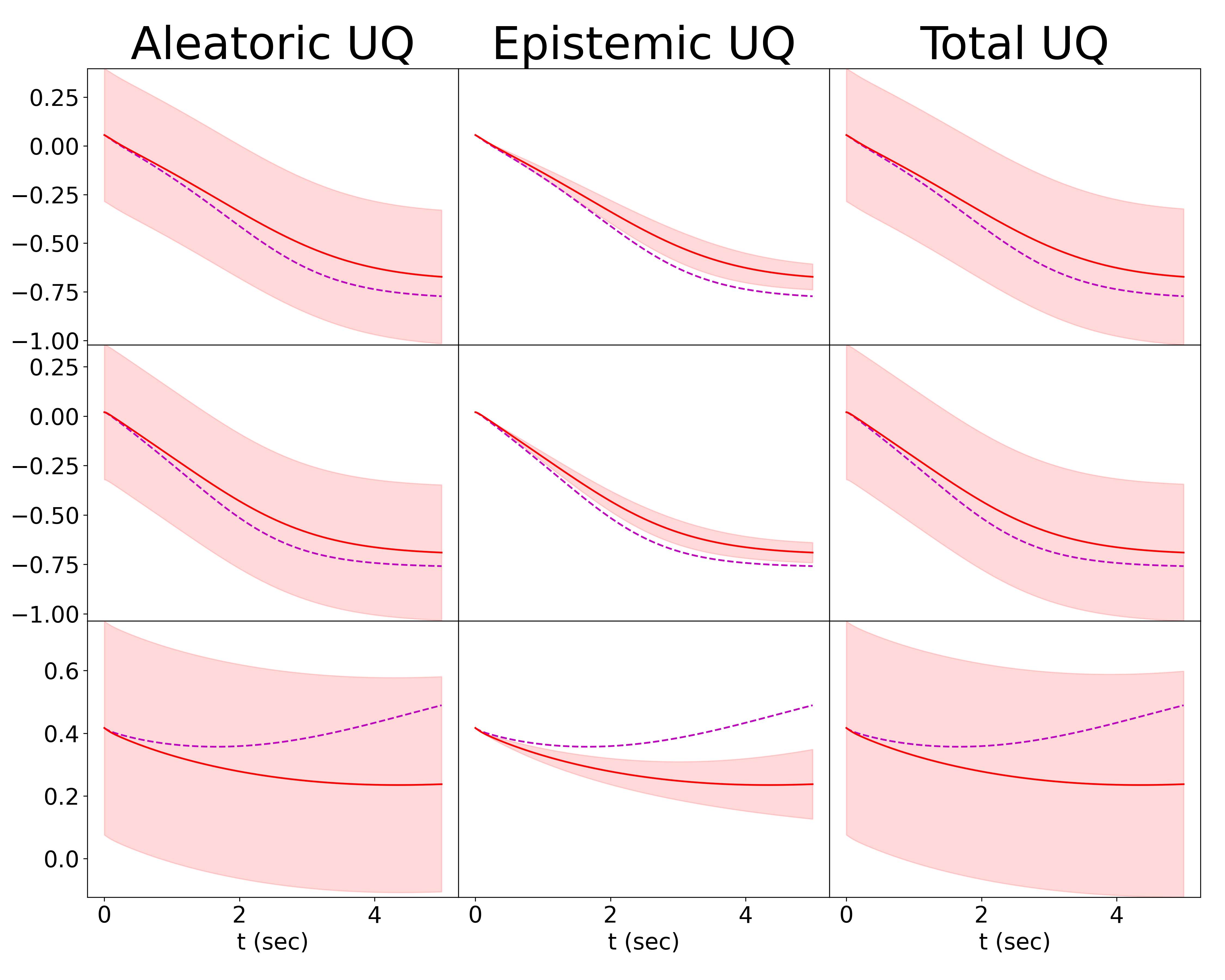}}
    \subfloat[$v_2$]{\includegraphics[width=0.5\textwidth]{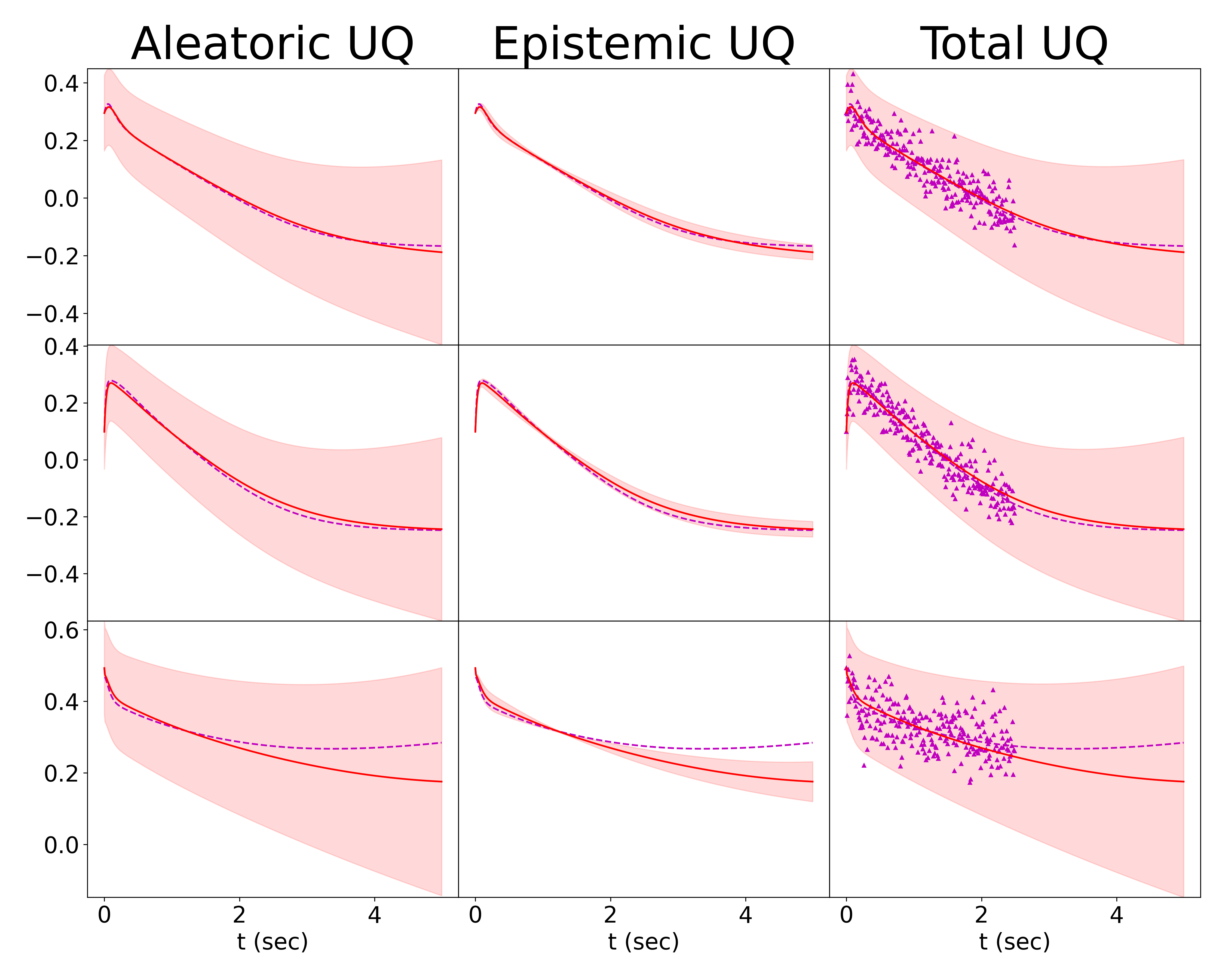}}
    \caption{Comparison of the DiffHybrid-UQ model’s prediction with UQ against the ground truth for $v_1$ (a) and $v_2$ (b) at three spatial locations over time. Training data: $\mathcal{D} = ({v}_2 \in \mathbb{R}^{20 \times 20} \times [0:0.01:3.5 s])$, with testing in $({v}_1 \in \mathbb{R}^{20 \times 20} \times [0:0.01:3.5 s], {v}_2 \in \mathbb{R}^{20 \times 20} \times (3.5:0.01:5 s])$.}
    \label{fig:pred_train_1D_rd_rnd}
\end{figure}

The inference capability of the DiffHybrid-UQ model is evaluated through an experimental study where the diffusion coefficients $D_i$ are assumed to be unknown and trainable. The model is trained on a single trajectory for up to 3.5 seconds ($\mathcal{D} = ({v}_2 \in \mathbb{R}^{20 \times 20} \times [0:0.01:3.5 s]) $) and then tested in the forecast region $(3.5:0.01:5 s]$. After 300 epochs of training, the outcomes are presented in Fig.~\ref{fig:pred_train_2D_rd_rnd} and Fig.~\ref{fig:pred_train_1D_rd_rnd}.
The results indicate strong prediction performance by the model, as the prediction mean agree with the ground truth well for both variables. As depicted in Fig.~\ref{fig:pred_train_2D_rd_rnd}, the predicted spatial uncertainty aligns well with expectations, which correlate with the prediction errors. The confidence interval shown in Fig.~\ref{fig:pred_train_1D_rd_rnd} effectively encapsulates the data scattering. Moreover, in line with prior observations, there is a notable increase in model-form uncertainty over time, further validating the model's effectiveness in dynamic modeling with UQ. 

\begin{figure}[!ht]
    \centering
    \includegraphics[width=0.6\linewidth]{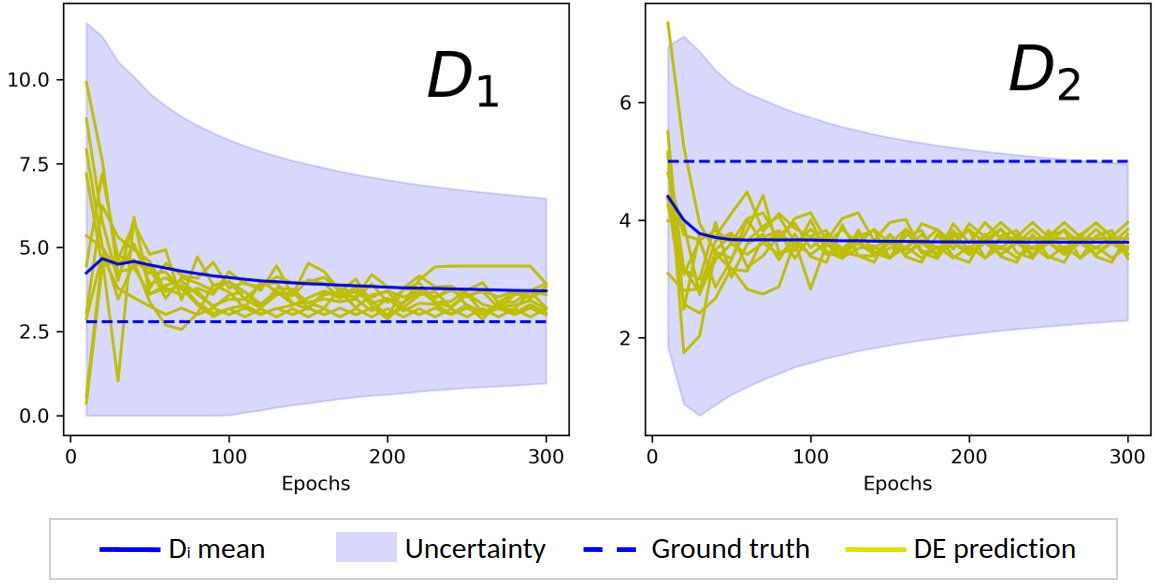}
    \caption{DiffHybrid-UQ inferred diffusion coefficients ($D_1$ and $D_2$) with quantified uncertainty over training epochs. DE represents ensembles of inferred diffusion coefficients.}
    \label{fig:Coeff_epoch_rd_gp}
\end{figure}

In this case, the model accurately predicts the dynamics of the unobserved state $v_1$ and also successfully infers the unknown diffusion coefficients $D_1$ and $D_2$ during the training process, as shown in Fig.~\ref{fig:Coeff_epoch_rd_gp}. Hybrid models are often susceptible to aliasing errors, particularly when trained with limited data. This issue can lead to inaccuracies in both the learned surrogate functions and the inferred coefficients, with the magnitude of errors typically increasing as the size of training data diminishes. In response to this challenge, the DiffHybrid-UQ model is designed to estimate errors in inferred parameters, thereby providing a more robust analysis. Figure~\ref{fig:Coeff_epoch_rd_gp} exemplifies the model's capability in quantifying uncertainty associated with the diffusion coefficients throughout the training process. This uncertainty, indicative of the aliasing effect, is observed to decrease as training advances. While it is not possible to guarantee absolute accuracy in the inferred diffusion coefficients due to the inverse and ill-posed nature of the problem, the DiffHybrid-UQ model enables a reasonably confident estimation. The fidelity of these inferences is contingent upon the dominant physics captured in the data, along with other influential factors, which will be elaborated upon in the Discussion section.

\begin{figure}[!ht]
    \centering
    \includegraphics[width=1\textwidth]{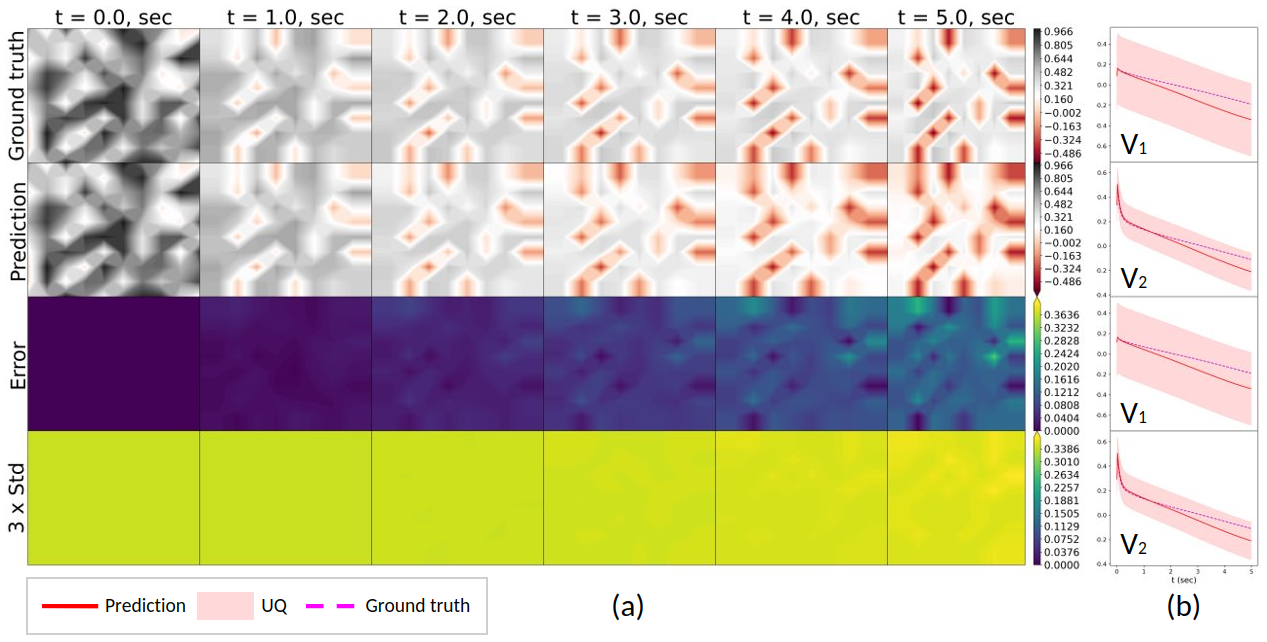}
    \caption{DiffHybrid-UQ testing results on an unseen randomly generated initial field, on a $10\times10$ grid with time-step of 0.02 sec.}
    \label{fig:pred_test_2D_rd_gp1}
\end{figure}
\begin{figure}[!ht]
    \centering
    \includegraphics[width=1\textwidth]{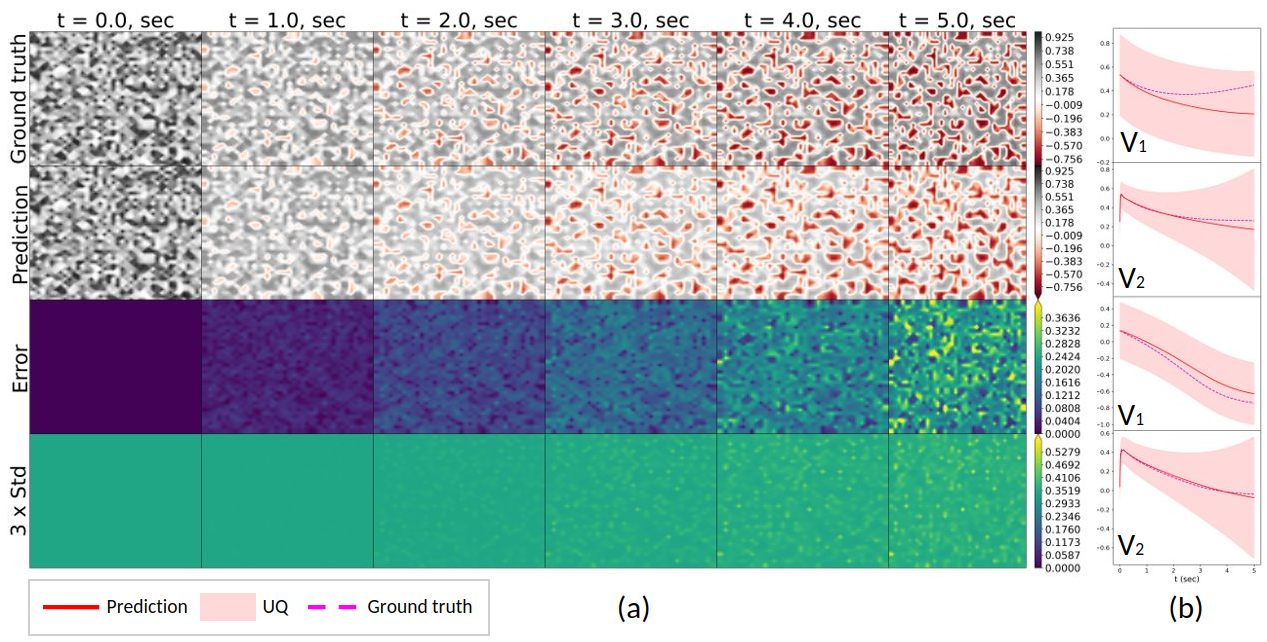}
    \caption{DiffHybrid-UQ testing results on an unseen randomly generated initial field, on a $40 \times40$ grid with time-step of 0.001 sec.}
    \label{fig:pred_test_2D_rd_gp2}
\end{figure}
To evaluate the DiffHybrid-UQ model's generalizability, we performed tests on various grid sizes and time steps, using a randomly generated initial conditions, which is distinct from the one used in the training phase. Figure~\ref{fig:pred_test_2D_rd_gp1} illustrates the test results on a $10 \times10$ grid with a time step of 0.02 seconds, while Fig.~\ref{fig:pred_test_2D_rd_gp2} presents the test results for a $40 \times40$ grid with a time step of 0.001 seconds. Across these tests, the model's predictions agree with the references reasonably well, demonstrating great generalizability. Notably, regions with higher deviation from the ground truth correspond to increased uncertainty in the model's predictions. This observation underscores the model's capability of quantifying uncertainty, even under varying testing conditions. It is crucial to highlight that, despite potential inaccuracies in the inferred diffusion coefficients, the trained surrogate model consistently demonstrates its ability to provide reliable predictions with quantified uncertainty.

\subsubsection{Correcting discretization errors and speedup}
In hybrid neural solvers, employing coarse spatio-temporal resolution is a common practice to enhance inference efficiency. However, this approach can introduce significant discretization errors in the PDE-integrated components, potentially affecting the accuracy of solutions. Within our DiffHybrid-UQ framework, this issue can be addressed by trainable neural network blocks, which are designed not only to learn and model unknown physical phenomena but also to concurrently correct for the discretization errors inherent in the coarser resolution. This dual capability allows the DiffHybrid-UQ model to maintain solution accuracy while benefiting from the computational efficiency of a coarser grid and larger timesteps. In this context, we explored the efficacy of the DiffHybrid-UQ model operating on coarse spatiotemporal resolutions. To this end, no structural modifications are needed, and the model's $\mathcal{U}_{nn}$ component serves a dual role: it acts as a surrogate for unknown functions and corrects solutions on coarser grids. In our evaluation, the model was trained on a coarse $20 \times20$ grid with a time step of 0.01 seconds, using data sampled from a high-resolution solution on a $200 \times 200$ grid with a finer time step of 0.0005 seconds. This approach allowed us to investigate the model's ability to replicate fine-grid precision on a coarser grid. The training data covered a single trajectory up to 3.7 seconds, denoted as $\mathcal{D} = (v_1, v_2 \in \mathbb{R}^{20 \times 20} \times [0:0.01:3.5 s] \sim \mathbb{R}^{200 \times 200} \times [0:0.0005:3.5 s])$,  and the model's predictions were subsequently assessed in the forecast period of 3.7 s to 5 s. 

\begin{figure}[!ht]
    \centering
    \includegraphics[width=0.9\textwidth]{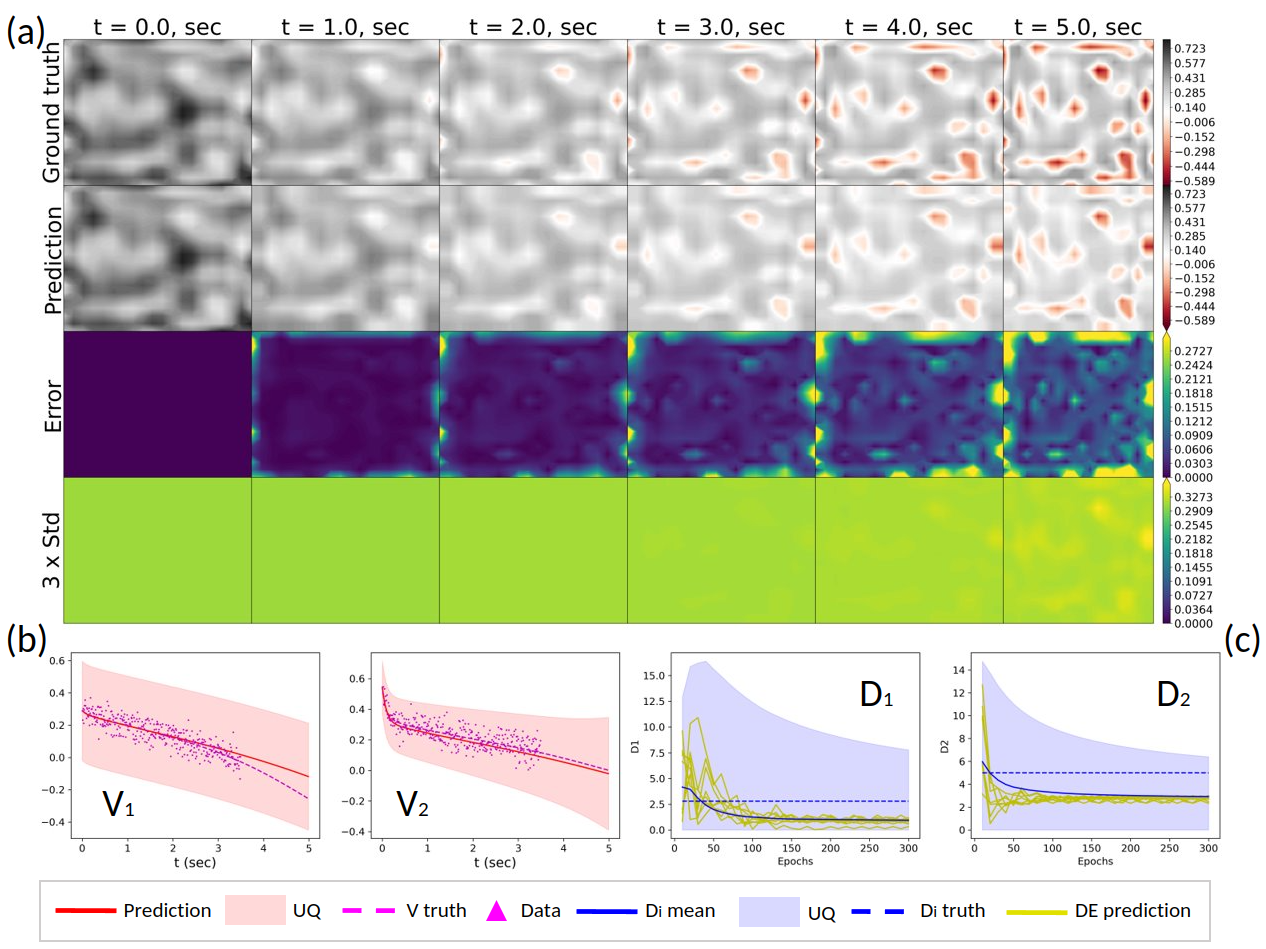}
    \caption{Compares the DiffHybrid-UQ model's predictions and UQ on a $20 \times20$ grid with a time step of 0.01 sec against the ground truth generated on a $200 \times200$ grid with a time step of 0.0005 sec. Training data spans $\mathcal{D} = (v_1, v_2 \in \mathbb{R}^{20 \times 20} \times [0:0.01:3.5 s] \sim \mathbb{R}^{200 \times 200} \times [0:0.0005:3.5 s])$ with testing in the $(v_1, v_2 \in \mathbb{R}^{20 \times 20} \times (3.5:0.01:5 s] \sim \mathbb{R}^{200 \times 200} \times (3.5:0.0005:5 s])$ region.}
    \label{fig:Pred_ND_GP_200x20_5e4t1e2_train2}
\end{figure}
Post-training for 500 epochs, the DiffHybrid-UQ model exhibited commendable performance within the training range. The results, presented in Fig.\ref{fig:Pred_ND_GP_200x20_5e4t1e2_train2}, showed reasonable prediction accuracy. However, a deviation from the ground truth was observed in the extrapolation region, where the model predicted a notably higher level of uncertainty, reflecting its recognition of increased error potential. This ability to predict with heightened uncertainty in less certain regions demonstrates the model's robustness, offering fine-grid level accuracy on coarser grids and thus reducing computational load. The efficiency of the DiffHybrid-UQ model, in terms of inference time and memory usage for both fine and coarse simulations, is detailed in Table.\ref{tab:perf}, highlighting its capability to produce high-fidelity results with lower computational demands and memory requirements.
\begin{table}[h!]
    \centering
    \begin{tabular}{ p{7cm} p{2.5cm} p{2.5cm} p{2.5cm}  }
     \hline
      Inference  & CPU time & GPU time & Memory\\
     \hline
     $\mathbb{R}^{2 \times 200 \times 200} \times [0:0.0005:5 s]$ &  0.67 sec & 0.66 sec & 33 GB \\
     $\mathbb{R}^{2 \times 20  \times 20 } \times [0:0.01:5 s]  $ &  8.8 sec  & 1.15 sec & 12 MB \\
     \hline
     \end{tabular}
     \caption{Inference time and memory consumption.}
     \label{tab:perf}
\end{table}

\subsection{Identifiability of unknown physical parameters}
The identifiability of physical parameters (e.g., diffusion coefficients) within the DiffHybrid-UQ framework, presents intriguing insights into the model's capability and limitations. As shown in cases presented above, a notable issue is the relatively low accuracy of inferred mean diffusion coefficients and associated high prediction uncertainty. This phenomenon can be attributed to the training data's dynamics being primarily driven by reaction physics, reducing the influence of the diffusion term on the overall dynamics. Therefore, the diffusion coefficients are less sensitive and difficult to be inferred accurately. We hypothesized that when the dynamics are more substantially influenced by the diffusion process, the model's ability to accurately infer diffusion coefficients would improve. 
\begin{figure}[!ht]
    \centering
    \includegraphics[width=0.9\textwidth]{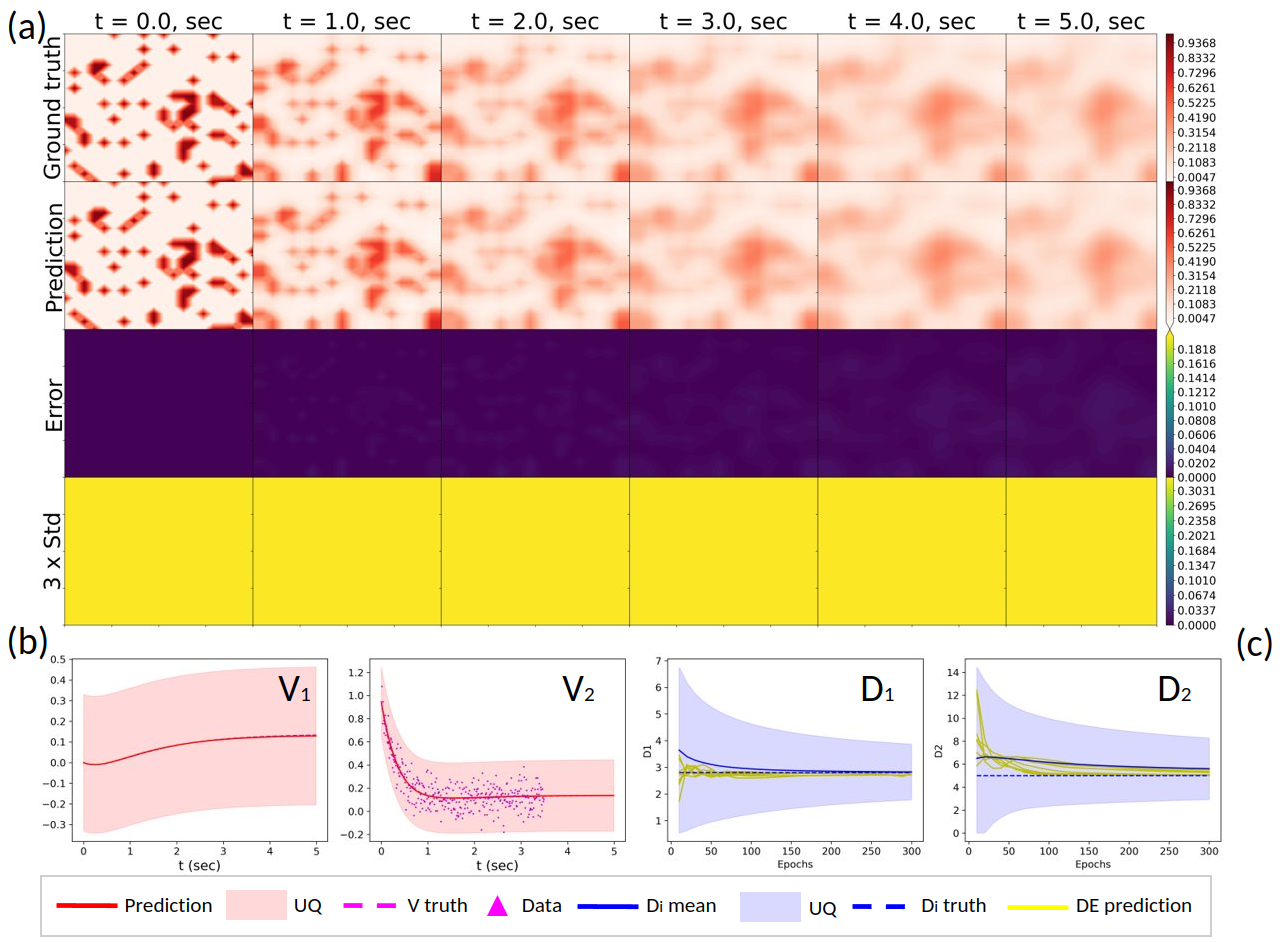}
    \caption{DiffHybrid-UQ test predictions with UQ, where diffusion coefficients are inferred simultaneously.} 
    \label{fig:DiffD}
\end{figure}
To test this hypothesis, a simulation with enhanced diffusion prominence was conducted, setting $D_1=2.8\times10^{-3}$ and $D_2=5.0\times10^{-3}$. This modification aimed to shift the focus towards the diffusion process. After 300 epochs of training, the results, as depicted in Fig.~\ref{fig:DiffD}, indicated a notable improvement in parameter inversion accuracy. The model's estimated confidence intervals closely captured the data/aleatoric uncertainty, and the inferred diffusion coefficients were more aligned with the ground truth. Furthermore, a decrease in model uncertainty was observed with each training epoch. These observations substantiate our hypothesis, emphasizing that the model's inferential accuracy is heavily dependent on the representation of governing physical processes in the training data. In scenarios where diffusion processes dominate, the model has enhanced capability in accurately inferring diffusion coefficients.

\section{Conclusion}
\label{sec:conclusion}

In this study, we have introduced DiffHybrid-UQ, a novel approximate Bayesian learning framework for quantifying uncertainty in physics-integrated hybrid neural differentiable models. This methodology, blending ensemble Bayesian learning with nonlinear transformations, has demonstrated substantial potential in effectively capturing and quantifying uncertainties associated with data-driven modeling of complex physical systems. Through rigorous evaluation on both ordinary and partial differential equations, the DiffHybrid-UQ model has demonstrated its proficiency in diverse aspects of UQ. Notably, it has accurately provided lower confidence levels for predictions in regions with sparse training data, effectively capturing the growth of uncertainty over time due to error accumulation. In scenarios involving parametric extrapolation, the model adeptly indicated increased uncertainty, particularly when initial conditions diverged from those in the training set. For spatiotemporal problems governed by PDEs, the DiffHybrid-UQ model can be trained with partial observational data, delivering reasonable predictions of spatiotemporal uncertainty. Its capacity to simultaneously infer physical parameters with quantified uncertainty, especially in the presence of aliasing errors, was particularly noteworthy. Furthermore, we also showcased the DiffHybrid-UQ model's ability to provide accurate predictions on coarse grids with reduced computational demands, underscoring its potential in practical applications where computational efficiency is critical. In sum, this work marks a crucial stride forward in the field of UQ for hybrid neural differentiable models that integrate physics-based numerical components with deep learning models using differentiable programming. This advancement holds great promise for refining physics-informed, data-driven decision-making processes across a wide array of scientific and engineering disciplines, fostering more accurate, robust, and trustworthy models for complex systems analysis.

Overall, this research represents a significant advancement in the field of uncertainty quantification for physics-integrated hybrid deep neural differentiable models. The successful integration of ensemble learning and nonlinear transformation provides a robust framework for enhancing prediction reliability and interpretability. The proposed method holds immense potential for improving hybrid simulation-based decision-making processes across various scientific and engineering domains.

\section*{Acknowledgment}
The authors would like to acknowledge the funds from the Air Force Office of Scientific Research (AFOSR), United States of America under award number FA9550-22-1-0065. JXW would also like to acknowledge the funding support from the Office of Naval Research under award number N00014-23-1-2071 and the National Science Foundation under award number OAC-2047127 in supporting this study.

\section*{Compliance with Ethical Standards}
Conflict of Interest: The authors declare that they have no conflict of interest.

\appendix

\section{Nomenclature}
\nomenclature{$\mathbf{v}$}{dynamic state variable}
\nomenclature{$\mathbf{v}_{\theta}$}{dynamic state variable approximated by the deterministic DiffHybrid model for a given $\boldsymbol{\theta}$}
\nomenclature{$\mathbf{V}_{\theta}$}{dynamic state variable distribution with mean $\mathbf{v}_{\theta}$, and covariance $ \text{diag}(\Sigma^2_{V,\theta})$ approximated by the probabilistic DiffHybrid-UQ model for a given $\boldsymbol{\theta}$}
\nomenclature{$\text{diag}(\Sigma^2_{V,\theta})$}{diagonal covariance matrix of state $\mathbf{v}$ approximated by the probabilistic DiffHybrid-UQ model for a given $\boldsymbol{\theta}$}
\nomenclature{$\mu_V$}{mean of state $\mathbf{v}$ approximated by the probabilistic DiffHybrid-UQ model considering all $\boldsymbol{\theta}$}
\nomenclature{$\text{diag}(\Sigma^2_{V})$}{diagonal covariance matrix of state $\mathbf{v}$ approximated by the probabilistic DiffHybrid-UQ modell considering all $\boldsymbol{\theta}$ }
\printnomenclature


\section{Stochastic Weight Averaging Gaussian (SWAG)}
\label{sec:SWAG}

The detailed SWAG~\cite{maddox2019simple} algorithm is described in this section. 
For a given SGD training trajectory $\boldsymbol{\theta}_{traj} = \{ \boldsymbol{\theta}_i \}_{i=1}^T = \{ \boldsymbol{\theta}_1, \boldsymbol{\theta}_2 , ... , \boldsymbol{\theta}_T \}$, the SWAG algorithm formulates an approximate Gaussian distribution $\mathcal{N}(\boldsymbol{\theta}_{\text{SWA}}, \frac{1}{2}(\Sigma_{\text{diag}} + \Sigma_{\text{low-rank}}))$ for posterior, where the mean $\boldsymbol{\theta}_{\text{SWA}}$ is calculated as
\begin{equation}
    \boldsymbol{\theta}_{\text{SWA}} = \bar{\boldsymbol{\theta}} = \frac{1}{T} \sum_{i=1}^T \boldsymbol{\theta}_i,
\end{equation}
and the diagonal covariance approximation $\Sigma_{\text{diag}}$ of the posterior distribution is calculated as
\begin{subequations}
    \begin{alignat}{2}
    \bar{\boldsymbol{\theta}^2} &= \frac{1}{T} \sum_{i=1}^T \boldsymbol{\theta}^2_i \\
    \Sigma_{\text{diag}} &= {diag}(\bar{\boldsymbol{\theta}^2} - \boldsymbol{\theta}^2_{\text{SWA}}).
    \end{alignat}
\end{subequations}
The sample covariance matrix of rank $T$ for a given SGD training trajectory $\boldsymbol{\theta}_{traj}$ can be calculated using Eq.~\ref{eq:sample_cov}. However, storing the whole trajectory $\boldsymbol{\theta}_{traj}$ for a model with a large number of parameters will be an infeasible task. Therefore, the SWAG method approximates the sample covariance with a low rank $r$ matrix as given in Eq.~\ref{eq:sample_cov_low_rank} and store deviation matrix $\hat{D}$ with columns $\textbf{d}_i = ({\boldsymbol{\theta}}_i - \bar{\boldsymbol{\theta}}_i)$.
\begin{subequations}
    \begin{alignat}{2}
    \Sigma &= \frac{1}{T-1} \sum_{i=1}^T (\boldsymbol{\theta}_i - \boldsymbol{\theta}_{\text{SWA}})(\boldsymbol{\theta}_i - \boldsymbol{\theta}_{\text{SWA}})^T
    \label{eq:sample_cov} \\
    \Sigma_{\text{low-rank}} &\approx \frac{1}{r-1} \sum_{i=T-r+1}^T ({\boldsymbol{\theta}}_i - \bar{\boldsymbol{\theta}}_i)(\boldsymbol{\theta}_i - \bar{\boldsymbol{\theta}}_i)^T = \frac{1}{r-1} \sum_{i=T-r+1}^T \hat{D}\hat{D}^T
    \label{eq:sample_cov_low_rank}
    \end{alignat}
\end{subequations}
Namely, instead of storing the whole trajectory, the posterior distribution statistics are updated while training according to the following algo~\cite{maddox2019simple}
\begin{subequations}
    \begin{alignat}{3}
        \bar{\boldsymbol{\theta}} &\leftarrow \frac{n\bar{\boldsymbol{\theta}} + \boldsymbol{\theta}_i}{n+1} \\
        \bar{\boldsymbol{\theta}^2} &\leftarrow \frac{n\bar{\boldsymbol{\theta}^2} + \boldsymbol{\theta}^2_i}{n+1} \\
        \hat{D} &\leftarrow \text{concatenate}( \hat{D}[:,2:], \boldsymbol{\theta}_i - \bar{\boldsymbol{\theta}} )
    \end{alignat}
\end{subequations}
and the following equation is utilized to obtain samples from $\mathcal{N}(\boldsymbol{\theta}_{\text{SWA}}, \frac{1}{2}(\Sigma_{\text{diag}} + \Sigma_{\text{low-rank}}))$ for prediction.
\begin{equation}
    \tilde{\boldsymbol{\theta}} = \boldsymbol{\theta}_{\text{SWA}} + \frac{1}{\sqrt{2}} \cdot \Sigma_{\text{diag}}^{1/2} z_1 + \frac{1}{\sqrt{2(r-1)}}\hat{D}z_2, \quad z_1 \sim \mathcal{N}(0,I_d), z_2 \sim \mathcal{N}(0, I_r)
\end{equation}

\section{Discretization of Governing PDE for the stochastic state $\mathbf{V}(\bar{\mathbf{x}})$}
\label{sec:Discretization}


Considering the probability distribution of all the random variables are diabolized Gaussian distribution, the mean and variance of the random variable, which is a linear combination of other random variables $\mathbf{Y} = \sum_i c_i \mathbf{X}_i$ is calculated as 
\begin{subequations}
    \begin{alignat}{2}
    \mu_Y = \sum_i c_i \mu_{Xi}
    \label{eq:PDE},\\
    \Sigma_Y^2 = \sum_i c_i^2 \Sigma_{Xi}^2
    \label{eq:BC},
    \end{alignat}
\end{subequations}

\subsection{Discretization in time}
Lets rewrite Eq.~\ref{eq:DiffHybrid-UQ-PDE}
$$\frac{\partial \mathbf{V}_{\boldsymbol{\theta}}(\bar{\mathbf{x}})}{\partial t} = \mathscr{F}_{nn}\big[\mathcal{K}(\mathbf{V}_{\boldsymbol{\theta}}(\bar{\mathbf{x}}); \boldsymbol{\lambda}_{\mathcal{K}}, \boldsymbol{\lambda}_{\mathcal{U}}), \mathcal{U}_{nn}(\mathbf{V}_{\boldsymbol{\theta}}(\bar{\mathbf{x}}); \boldsymbol{\lambda}_{\mathcal{K}}, \boldsymbol{\lambda}_{\mathcal{U}},\boldsymbol{\theta}_{nn}); \boldsymbol{\theta}_{nn}\big]$$
as
\begin{equation}
    \frac{\partial }{\partial t} (\mathbf{V}({\mathbf{x}}, t) | \boldsymbol{\theta}) = \mathscr{F}_{nn}(t, \mathbf{V} ({\mathbf{x}}, t) | \boldsymbol{\theta}).
    \label{eq:13a}
\end{equation}

Now to discretize Eq.~\ref{eq:13a} in time, we approximate the spatiotemporal random state variable $\mathbf{V}({\mathbf{x}}, t)$ as a spatial random state variable at various times $\{0,...,t,...,T_r\}$. 
$$\mathbf{V}({\mathbf{x}}, t) \approx \{\mathbf{V}^{(t)}({\mathbf{x}})\}_{t=0}^{T_r} $$

Equation.~\ref{eq:13a} can be discretized in time using the Euler method as
\begin{equation}
    (\mathbf{V}^{(t+1)}({\mathbf{x}})| \boldsymbol{\theta}) = (\mathbf{V}^{(t)}({\mathbf{x}})| \boldsymbol{\theta}) + \Delta t  \mathscr{F}_{nn} (t,\mathbf{V}^{(t)}({\mathbf{x}})| \boldsymbol{\theta}),
\end{equation}
where the mean and the variance of the Gaussian random variable \textbf{V} will be updated as
\begin{subequations}
    \begin{alignat}{3}
    (\mu_{K1}, \Sigma^2_{K1}) &=(\mu_{\mathscr{F}_{nn}}, \Sigma^2_{\mathscr{F}_{nn}}) = \mathscr{F}_{nn} \big(t, (\mu_{V^t}, \Sigma^2_{V^t})| \boldsymbol{\theta} \big),\\
    (\mu_{V^{t+1}}|\mu_{V^{t}}, \boldsymbol{\theta}) &= \mu_{V^t} + \Delta t  \mu_{K1} \\
    (\Sigma^2_{V^{t+1}}|\Sigma^2_{V^{t}}, \boldsymbol{\theta}) &= \Sigma_{V^t}^2 + \Delta t^2  \Sigma^2_{K1}.
    \label{eq:BC},
    \end{alignat}
\end{subequations}

Similarly, eq.~\ref{eq:13a} can be discretized in time using the Runge-Kutta 4th-order method as
\begin{subequations}
    \begin{alignat}{6}
    (\mu_{K1}, \Sigma^2_{K1}) &= \mathscr{F}_{nn} \big(t, (\mu_{V^t}, \Sigma^2_{V^t})| \boldsymbol{\theta} \big) , \\
    (\mu_{K2}, \Sigma^2_{K2}) &= \mathscr{F}_{nn} \big(t + 0.5\Delta t, (\mu_{V^t} + 0.5\Delta t \mu_{K1}, \Sigma^2_{V^t} + 0.25\Delta t^2 \Sigma^2_{K1})| \boldsymbol{\theta} \big) , \\
    (\mu_{K3}, \Sigma^2_{K3}) &= \mathscr{F}_{nn} \big(t + 0.5\Delta t, (\mu_{V^t} + 0.5\Delta t \mu_{K2}, \Sigma^2_{V^t} + 0.25\Delta t^2 \Sigma^2_{K2})| \boldsymbol{\theta} \big) , \\
    (\mu_{K4}, \Sigma^2_{K4}) &= \mathscr{F}_{nn} \big(t +    \Delta t, (\mu_{V^t} +    \Delta t \mu_{K3}, \Sigma^2_{V^t} +     \Delta t^2 \Sigma^2_{K3})| \boldsymbol{\theta} \big) , \\
    (\mu_{V^{t+1}}|\mu_{V^{t}}, \boldsymbol{\theta})  &= \mu_{V^t} + \frac{\Delta t}{6} \big( \mu_{K1} + 2\mu_{K2} + 2\mu_{K3} + \mu_{K4} \big)\\
    (\Sigma^2_{V^{t+1}}|\Sigma^2_{V^{t}}, \boldsymbol{\theta})  &= \Sigma^2_{V^t} + \frac{\Delta t^2}{36} \big( \Sigma^2_{K1} + 4\Sigma^2_{K2} + 4\Sigma^2_{K3} + \Sigma^2_{K4} \big)
    \end{alignat}
\end{subequations}
where $\mathbf{K}_i$ are Gaussian random variables $\mathbf{K}_i(\mathbf{x} | \boldsymbol{\theta}) \sim \mathcal{N}(\mu_{Ki}(\mathbf{x}, t|\boldsymbol{\theta}), \text{diag}(\Sigma^2_{Ki}(\mathbf{x}, t|\boldsymbol{\theta})))$.

\subsection{Discretization in space}

Now to discretize the right-hand side of Eq.~\ref{eq:13a} in space, we approximate the spatial random state variable $\mathbf{V}^{(t)}({\mathbf{x}})$ as a random state variable at a grid point $x$. 
$$\mathbf{V}^{(t)}({\mathbf{x}}) \approx \{\mathbf{V}^{(t)}_{(x)}\}_{x=0}^{n_{grid}}. $$
Any method can be used to discretize the above equation in space. Here in this study, FDM is used. 
The discretization results in Gaussian random vector $\mathbf{V}({\mathbf{x}}, t) \approx \{\mathbf{V}^{(t)}({\mathbf{x}})\}_{t=0}^{T_r} \approx \{\{\mathbf{V}^{(t)}_{(x)}\}_{x=0}^{n_{grid}}\}_{t=0}^{T_r}$. And for 2D, $\mathbf{V} \sim \mathcal{N}(\mathbf{V}| \mu_{V}, \text{diag}(\Sigma_{V}^2))$, where $\mu_{V} \in \mathbb{R}^{n_t \times n_v \times (n_x \times n_y)}$ and $\text{diag}(\Sigma_{V}^2) \in \mathbb{R}^{n_t \times n_v \times (n_x \times n_y)}$ are the mean and co-variance vector. Here $n_t$, $n_v$, $n_x$, and $n_y$ are the number of time steps, the number of variables, and the number of grid points in the x and y directions, respectively.

For instance, for given  $\mathbf{V}^{(t)} \sim \mathcal{N}(\mathbf{V}^{(t)}| \mu^{(t)}, \text{diag}({\Sigma^2}^{(t)}))$, where $\mu^{(t)} \in \mathbb{R}^{n_v \times (n_x \times n_y)}$ and $\text{diag}({\Sigma^2}^{(t)}) \in \mathbb{R}^{n_v \times (n_x \times n_y)}$ are the mean and co-variance vector, the discretization of the Laplace term results in
$$ \nabla^2 V^{(t)}_{(i,j)} = \frac{V^{(t)}_{(i+1,j)}+V^{(t)}_{(i-1,j)}-2V^{(t)}_{(i,j)}}{\Delta x^2} + \frac{V^{(t)}_{(i,j+1)}+V^{(t)}_{(i,j-1)}-2V^{(t)}_{(i,j)}}{\Delta y^2}$$
And as the Laplace is a linear operator, we can write 
\begin{align}
\nabla^2 \mu^{(t)}_{(i,j)} &= \frac{\mu^{(t)}_{(i+1,j)}+\mu^{(t)}_{(i-1,j)}-2\mu^{(t)}_{(i,j)}}{\Delta x^2} \nonumber \\
&\quad + \frac{\mu^{(t)}_{(i,j+1)}+\mu^{(t)}_{(i,j-1)}-2\mu^{(t)}_{(i,j)}}{\Delta y^2} \\
\nabla^2 \text{diag}({\Sigma^2}^{(t)}_{(i,j)}) &= \frac{\text{diag}({\Sigma^2}^{(t)}_{(i+1,j)})+\text{diag}({\Sigma^2}^{(t)}_{(i-1,j)})+4\text{diag}({\Sigma^2}^{(t)}_{(i,j)})}{\Delta x^4} \nonumber \\
&\quad + \frac{\text{diag}({\Sigma^2}^{(t)}_{(i,j+1)})+\text{diag}({\Sigma^2}^{(t)}_{(i,j-1)})+4\text{diag}({\Sigma^2}^{(t)}_{(i,j)})}{\Delta y^4}.
\end{align}
where $i, j$ are x , y indices.

\section{Uncertainty Propagation through functions}

\subsection{Uncertainty Propagation through known functions using Unscented-Transform method}
\label{sec:UTfun}

As discussed in the methodology section, the known non-linear operators $\mathcal{K}$ have been designed to handle probability distribution with the help of the UT method. This is achieved by wrapping the deterministic operator $\mathcal{K}_d$ with the UT function. Detailed procedure is given below.

The UT function takes the mean and variance of the input dynamic variable $\mathbf{V}$ and calculates sigma points $\mathcal{X}$ using the following formulation\cite{wan2000unscented}.

$$
\begin{array}{ll}
\mathcal{X}_0 & = \mu_V \\
\mathcal{X}_i & = \mu_V + \left(\sqrt{(L+\lambda) \Sigma^2_V }\right)_i \quad i=1, \ldots, L \\
\mathcal{X}_i & = \mu_V - \left(\sqrt{(L+\lambda)\Sigma^2_V }\right)_{i-L} \quad i=L+1, \ldots, 2 L \\
W_0^{(m)} & =\lambda /(L+\lambda) \\
W_0^{(c)} & =\lambda /(L+\lambda)+\left(1-\alpha^2+\beta\right) \\
W_i^{(m)} & =W_i^{(c)}=1 /\{2(L+\lambda)\} \quad i=1, \ldots, 2 L
\end{array}
$$
where $\lambda=\alpha^2(L+\kappa)-L$ is a scaling parameter. $L$ is the dimension of the dynamic random variable $V$, $\alpha$ determines the spread of the sigma points around $\mu_V$, $\kappa$ is a secondary parameter, and $\beta$ is used to incorporate prior knowledge of of $\mathbf{V}$ distribution. Here for current study we used $\alpha = 1$, $\kappa = 0$, and $\beta = 2$.
These sigma vectors are thereafter propagated through the nonlinear function $\mathcal{K}_d$,
$$
\mathcal{Y}_i=\mathcal{K}_d \left(\mathcal{X}_i\right) \quad i=0, \ldots, 2 L
$$
and the mean and covariance for $\mathbf{y}$ are approximated using a weighted sample mean and covariance of the posterior sigma points,
$$
\begin{aligned}
\mu_y & \approx \sum_{i=0}^{2 L} W_i^{(m)} \mathcal{Y}_i \\
\Sigma^2_{\mathbf{y}} & \approx \sum_{i=0}^{2 L} W_i^{(c)}\left\{\mathcal{Y}_i-\mu_y\right\}\left\{\mathcal{Y}_i-\mu_y\right\}^T.
\end{aligned}
$$
Detailed algorithm for operator $\mathcal{K}$ is given in algo.~\ref{alg:UT}.

\begin{algorithm}[hbt!]
{
\caption{An algorithm for uncertainty propagation through nonlinear function}\label{alg:UT}
  \SetKwFunction{FSum}{$\mathcal{K}$}
 
  \SetKwProg{Fn}{Function}{:}{}
  \Fn{\FSum{$\mathcal{K}_d, \mathbf{V}=(\mu_V, \Sigma^2_V), \alpha, \beta, \kappa$}}{
        $L \gets dim(\mu_V)$ \\
        $\lambda \gets \alpha^2(L+\kappa)-L$ \\
        $W_0^{(m)} \gets \lambda/(L+\lambda) $ \Comment {Calculate weights}\\
        $W_0^{(c)} \gets \lambda/(L+\lambda) + (1-\alpha^2 + \beta) $ \\
        $\mathcal{X}_0 \gets \mu_V$ \Comment {Calculate Sigma points}\\
        \For{i = 1 to $L$}
        {
         $W_i^{(m)} \gets 1/2(L+\lambda) $ \Comment {Calculate weights}\\
         $W_i^{(c)} \gets 1/2(L+\lambda) $\\
         $\mathcal{X}_i \gets \mu_V + (\sqrt{(L+\lambda)\Sigma^2_V})_i$ \Comment {Calculate Sigma points}\\
         $\mathcal{X}_{i+L} \gets \mu_V - (\sqrt{(L+\lambda)\Sigma^2_V})_i$ \\
         $\mathcal{Y}_i \gets \mathcal{K}_d(\mathcal{X}_i)$ \Comment {Propagate $\mathcal{X}$ through $\mathcal{K}_n$} \\
         $\mathcal{Y}_{i+L} \gets \mathcal{K}_d(\mathcal{X}_{i+L})$ 
        }
         $\mu_y \gets \sum_{i=0}^{2L} W_i{(m)} \mathcal{Y}_i $ \Comment {Calculate mean and variance of o/p} \\
         $\Sigma^2_y \gets \sum_{i=0}^{2L} W_i{(c)} \{\mathcal{Y}_i-\mu_y\}\{\mathcal{Y}_i-\mu_y\}^T$ 
        
        \KwRet $\mu_y, \Sigma^2_y$\ \Comment {Output random variable}
  }}
\end{algorithm}

 
        
        

\subsection{Uncertainty Propagation through Neural-Network}
\label{sec:NNfu}

\begin{figure}[!ht]
\centering
\includegraphics[width = 1.0\textwidth]{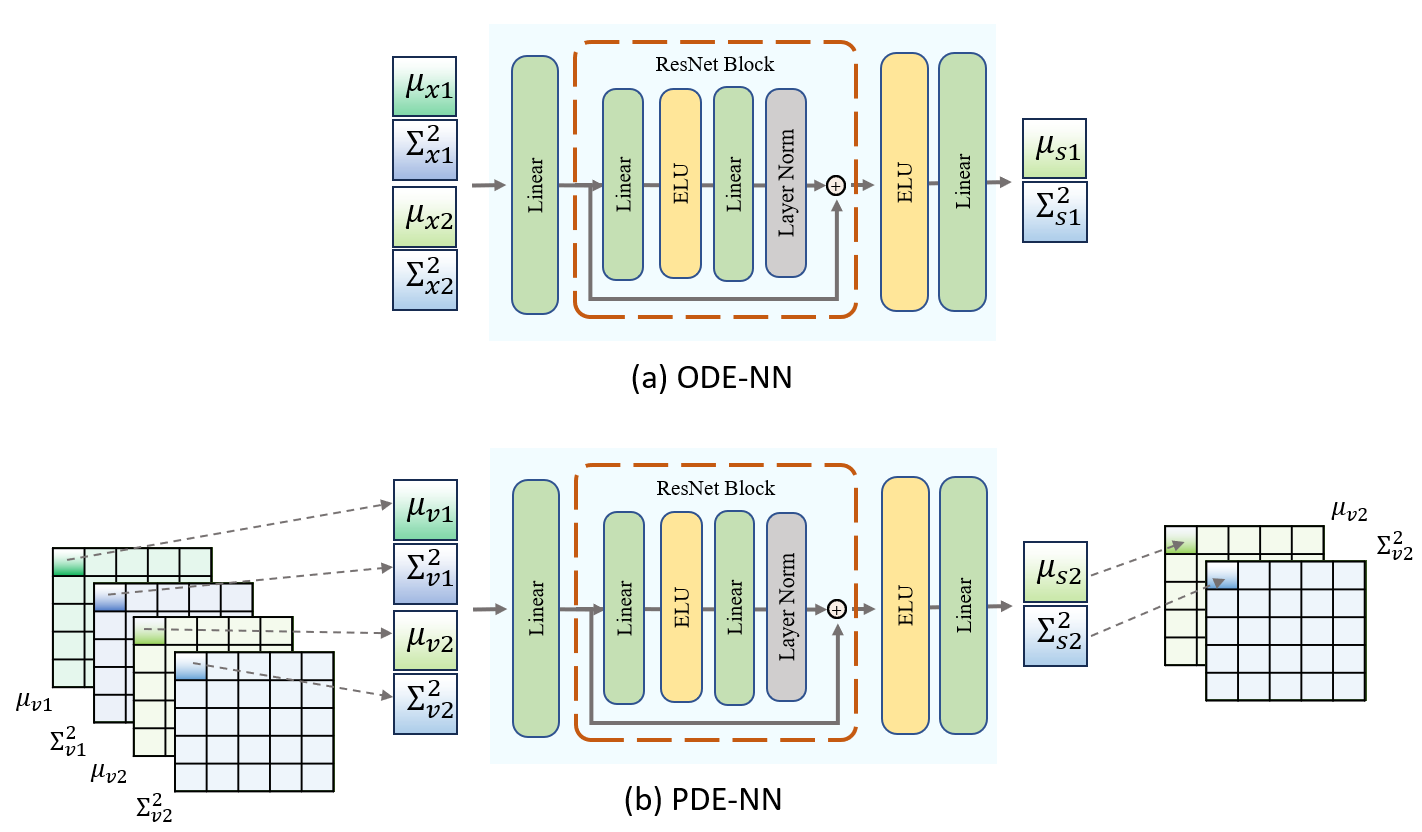}
\caption{Neural network architecture used in DiffHybrid-UQ model.}
\label{fig:NN-architecture}
\end{figure}

Figure~\ref{fig:NN-architecture} illustrates the Neural Network architectures utilized for learning the unknown functions in both ODE and PDE systems. For ODE-NN, each hidden layer has 8 neurons, while in the case of PDE-NN, 32 neurons were employed. The PDE-NN network follows a point-to-point approach, taking independent distribution statistics (first- and second-order moments) inputs at every point of the grid and generating the corresponding output distribution statistics at that specific grid point. 

\section{Algorithms}

The detailed algorithms used to construct a DiffHybrid-UQ solver for Hamiltonian ODEs and Reaction-Diffusin PDEs case are shown in Alog.~\ref{alg:ODE} and Alog.~\ref{alg:PDE}, respectively.

\begin{algorithm}[hbt!]
{
\caption{An algorithm for DiffHybrid-UQ solver for Hamiltonian ODEs system}\label{alg:ODE}
  \SetKwFunction{FSum}{DiffHybrid-UQ}
 
  \SetKwProg{Fn}{Function}{:}{}
  \Fn{\FSum{$\mathbf{x}_{init}=(\mu^0_x, \Sigma^0_x),  \boldsymbol{\theta} = (\boldsymbol{\theta}_{\lambda_{U}}, \boldsymbol{\theta}_{\mathcal{NN}}, \boldsymbol{\theta}_{\Sigma^2_0})$}}{
        $t \gets 0$ \\
        \While{$t < t\_max$} 
        { 
            $(\mu_{s_1}, \Sigma^2_{s_1})  \gets \mathcal{NN}^r_{\mathcal{U}}((\mu_{x_1}, \Sigma^2_{x_1})^t, \boldsymbol{\theta})$ \Comment {Compute source with UT}\\
            $(\mu_{s_2}, \Sigma^2_{s_2})  \gets  \mathcal{K}^r_n((\mu_{x_2}, \Sigma_{x_2})^t, \boldsymbol{\theta})$ \Comment {Compute source with NN}\\
            \Comment {Euler time stepping, (can use RK4)}\\
            $(\mu_{x_i}, \Sigma^2_{x_i})^{t+1}  \gets (\mu_{x_i}, \Sigma^2_{x_i})^t +  (\Delta t \mu_{s_i}, \Delta t^2 \Sigma^2_{s_i})$ 
        }
        \KwRet $(\{\mu_{x_i}\}_{t=0}^{t\_max}, \{\Sigma^2_{x_i}\}_{t=0}^{t\_max})$\ \Comment {Predicted time series}
  }}
\end{algorithm}

\begin{algorithm}[hbt!]
{
\caption{An algorithm for DiffHybrid-UQ solver for reaction-diffusion PDEs system}\label{alg:PDE}
  \SetKwFunction{FSum}{DiffHybrid-UQ}
 
  \SetKwProg{Fn}{Function}{:}{}
  \Fn{\FSum{$\mathbf{V}_{init}=(\mu^0_V, \Sigma^0_V),  \boldsymbol{\theta} = (\boldsymbol{\theta}_{\lambda_{U}}, \boldsymbol{\theta}_{\mathcal{NN}}, \boldsymbol{\theta}_{\Sigma^2_0})$}}{
        $t \gets 0$ \\
        \While{$t < t\_max$} 
        { 
            $(\mu_{\nabla^2V_i}, \Sigma^2_{\nabla^2V_i}) \gets \text{Laplace}((\mu^t_{V_i}, \Sigma^t_{V_i}), \boldsymbol{\theta})$ \Comment {Construct Laplace}\\
            $(\mu_{s_1}, \Sigma^2_{s_1})  \gets  \mathcal{K}^r_n((\mu_{V_1}, \Sigma_{V_1})^t, \boldsymbol{\theta})$ \Comment {Compute source with UT}\\
            $(\mu_{s_2}, \Sigma^2_{s_2})  \gets  \mathcal{NN}^r_{\mathcal{U}}((\mu_{V_2}, \Sigma_{V_2})^t, \boldsymbol{\theta})$ \Comment {Compute source with NN}\\
            \Comment {Euler time stepping, (can use RK4)}\\
            $(\mu_{V_i}, \Sigma^2_{V_i})^{t+1}  \gets (\mu_{V_i}, \Sigma^2_{V_i})^t +  (\Delta t (\boldsymbol{\theta}_{D_i} \mu_{\nabla^2V_i} + \mu_{s_i}), \Delta t^2 (\boldsymbol{\theta}^2_{D_i} \Sigma^2_{\nabla^2V_i} + \Sigma^2_{s_i}))$ \\
            $(\mu_{V_i}, \Sigma^2_{V_i})^{t+1} \gets \text{Boundary\_Condition}((\mu_{V_i}, \Sigma^2_{V_i})^{t+1})$
        }
        \KwRet $(\{\mu_{V_i}\}_{t=0}^{t\_max}, \{\Sigma^2_{V_i}\}_{t=0}^{t\_max})$\ \Comment {Predicted time series}
  }}
\end{algorithm}

\section{Additional Computational Results}

\subsection{Hamiltonian }

\begin{figure}[!ht]
    \centering
    \subfloat{\includegraphics[width=0.8\linewidth]{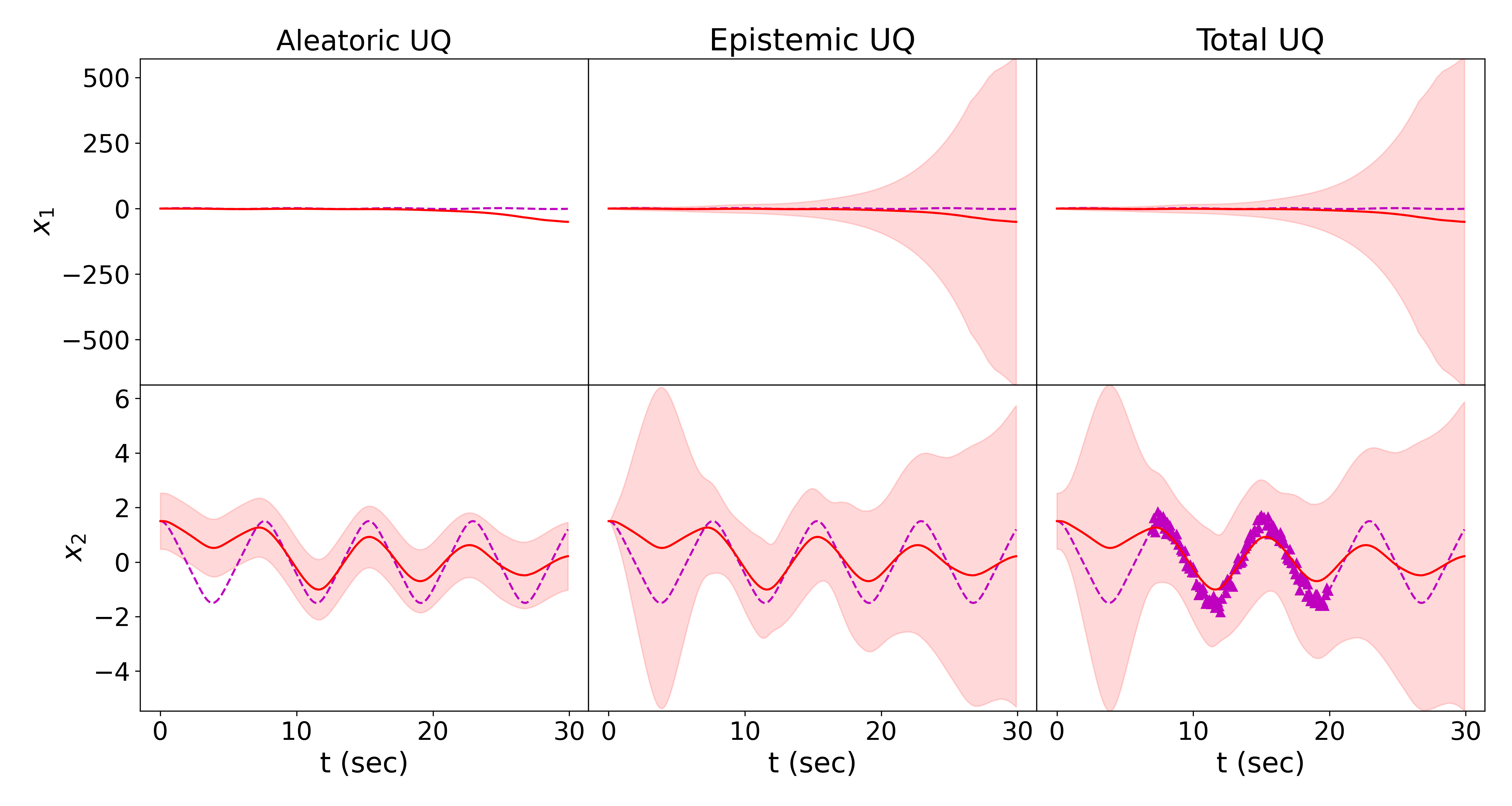}}\\
    \subfloat{\includegraphics[width=0.6\textwidth]{./fig/label_pugd.png}}
    \caption{Comparison of the DiffHybrid-UQ model’s prediction with UQ against the ground truth.
    Training data is $\mathcal{D} = (x_2 \in \mathbb{R}^1 \times [5:0.1:20 s])$, with 
    testing in $(x_1 \in \mathbb{R}^1 \times [0:0.1:30 s], x_2 \in \mathbb{R}^1 \times [0:0.1:5 s) \cup (20:0.1:30 s])$ region. }
    \label{fig:pred_C3_Ham}
\end{figure}

In addition to the test cases shown in the Result section, one more case was conducted on the Hamiltonian ODEs system, where the DiffHybrid-UQ model is trained by only providing the data for the $x_2$ variables in 5 to 20 sec, specifically $\mathcal{D} = ({x}_2 \in \mathbb{R}^1 \times [5:0.1:20s]) = \mathbb{R}^1 \times [0:0.1:20s]$. The training process involved 10,000 epochs, and the results obtained are presented in Fig.~\ref{fig:pred_C3_Ham}.

Due to the limited amount of provided data, the predicted uncertainty is significantly large, except in the region where the data is available. Despite these challenging conditions, the predicted model uncertainty exhibits reasonable behavior. For instance, in the 0 to 5-second region, the model uncertainty for $x_2$ remains high, while for $x_1$, it increases rapidly.

\subsection{Reaction-Diffusion}

\begin{figure}[!ht]
    \centering
    \includegraphics[width=0.9\textwidth]{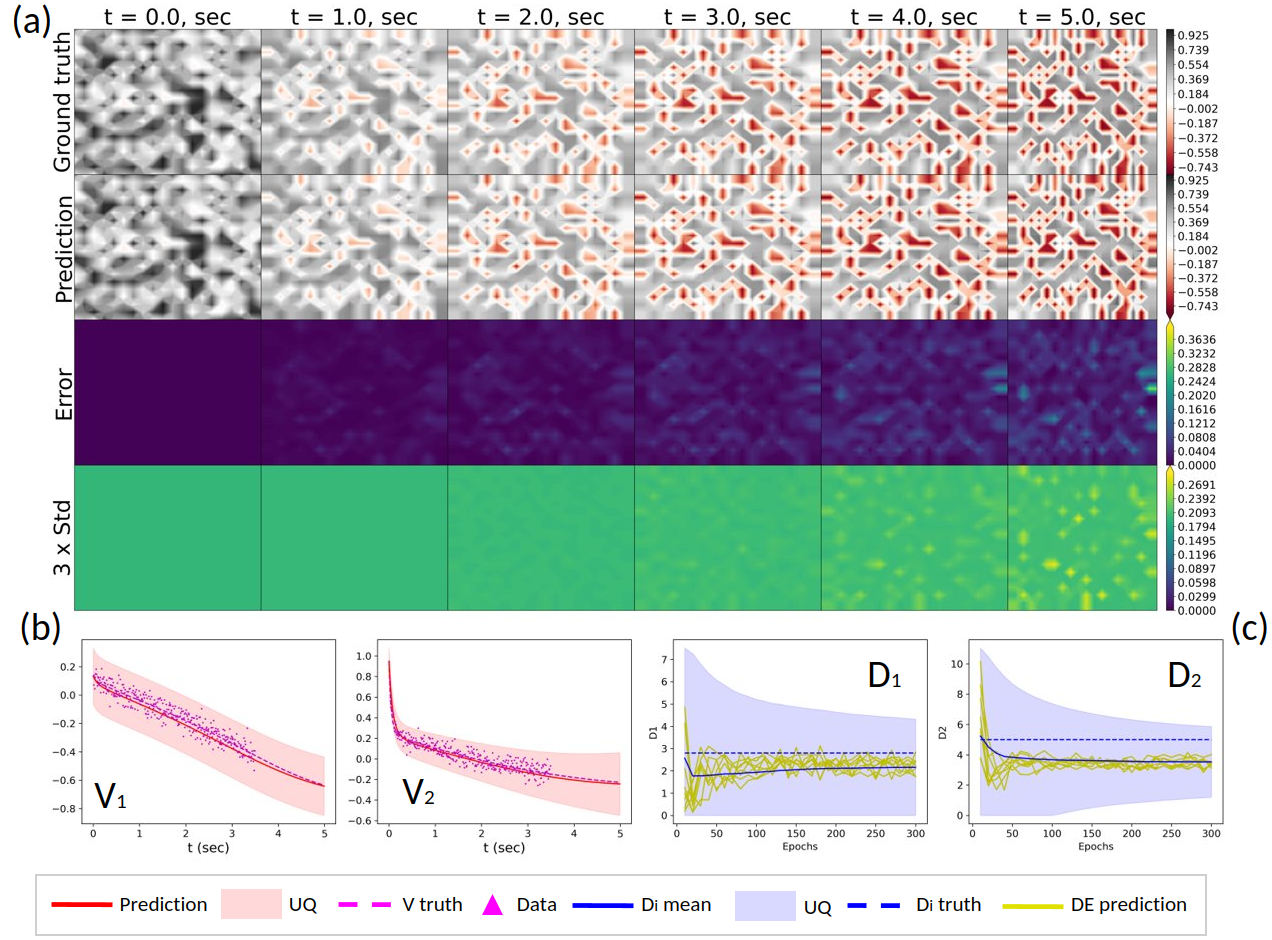}
    \caption{Compares the DiffHybrid-UQ model's predictions and UQ on a $20 \times20$ grid with a time step of 0.01 sec against the ground truth. Training data spans $\mathcal{D} = (v_1, v_2 \in \mathbb{R}^{20 \times 20} \times [0:0.01:3.5 s])$ with testing in the $(v_1, v_2 \in \mathbb{R}^{20 \times 20} \times (3.5:0.01:5 s])$ region.}
    \label{fig:pred_2D_N_GP_train2}
\end{figure}

Furthermore, in addition to the test cases discussed in the Results section, we present two additional test cases involving the Reaction-Diffusion PDEs system. Figure~\ref{fig:pred_2D_N_GP_train2} displays the results obtained after training the model for 300 epochs on a $20 \times 20$ grid with a time-step of 0.01 sec. In this case, data for $v_1$ and $v_2$ is provided up to 3.5 sec, represented as $\mathcal{D} = (v_1, v_2 \in \mathbb{R}^{20 \times20} \times [0:0.01:3.5 s]) = \mathbb{R}^{2 \times20 \times20} \times [0:0.01:3.5 s]$.
Comparing this case to a scenario where only data for $v_2$ is provided (Fig.~\ref{fig:pred_train_2D_rd_rnd} and Fig.~\ref{fig:pred_train_1D_rd_rnd}), we observe a reduction in error, improved accuracy in inferring the diffusion coefficients, and accordingly reduction in confidence interval.

\begin{figure}[!ht]
    \centering
    \includegraphics[width=0.9\textwidth]{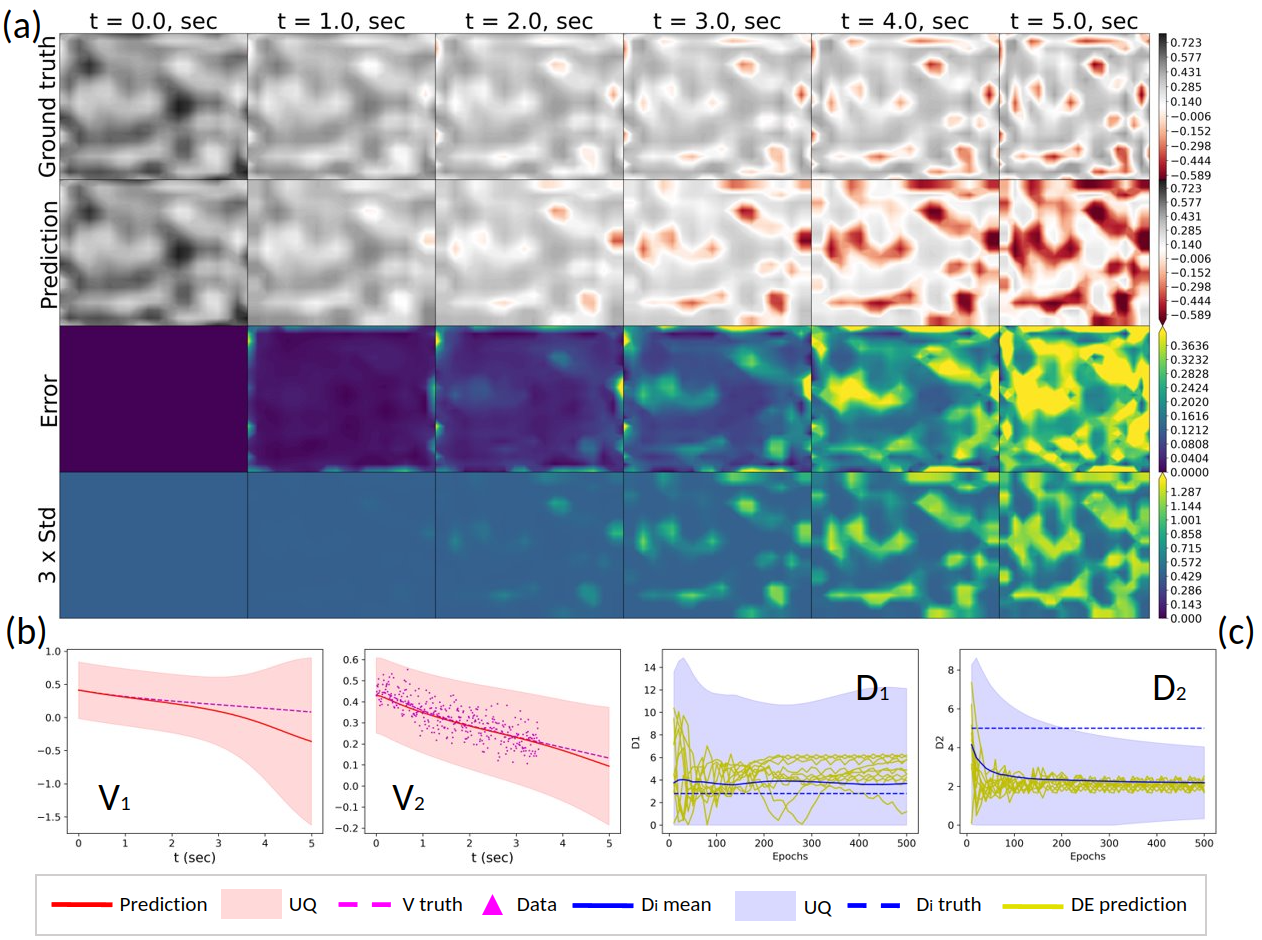}
    \caption{Compares the DiffHybrid-UQ model's predictions and UQ on a $20 \times20$ grid with a time step of 0.01 sec against the ground truth generated on a $200 \times200$ grid with a time step of 0.0005 sec. Training data spans $\mathcal{D} = (v_2 \in \mathbb{R}^{20 \times 20} \times [0:0.01:3.5 s] \sim \mathbb{R}^{200 \times 200} \times [0:0.0005:3.5 s])$ with testing in the $(v_1 \in \mathbb{R}^{20 \times 20} \times [0:0.01:5 s] \sim \mathbb{R}^{200 \times 200} \times [0:0.0005:5 s], v_2 \in \mathbb{R}^{20 \times 20} \times (3.5:0.01:5 s] \sim \mathbb{R}^{200 \times 200} \times (3.5:0.0005:5 s])$ region.}
    \label{fig:Pred_ND_GP_200x20_5e4t1e2_train}
\end{figure}

Similarly to the test case described in the Corrector Network section, we present the results in Figure~\ref{fig:Pred_ND_GP_200x20_5e4t1e2_train}, where the training data is provided solely for $v_2$. The model is trained for 500 epochs. In the training region, which extends up to 3.5 sec, the results exhibit reasonable accuracy. However, as we move into the extrapolation region for $v_1$, the predictions deviate from the ground truth. Notably, the predicted uncertainty in the extrapolation region is also high, indicating the lack of reliability in the Hybrid model's prediction.




\bibliographystyle{elsarticle-num}
\bibliography{ref,myRef}

\end{document}